\documentclass{article}

     \PassOptionsToPackage{numbers, compress}{natbib} 

    \usepackage[preprint]{neurips_2022}

\usepackage[utf8]{inputenc} 
\usepackage[T1]{fontenc}    
\usepackage{hyperref}       
\usepackage{url}            
\usepackage{booktabs}       
\usepackage{nicefrac}       
\usepackage{microtype}      
\usepackage{xcolor}         
\usepackage{hyperref}
\usepackage{amsmath,amssymb,amsfonts,mathtools}
\usepackage{bm}
\usepackage{subfigure}
\usepackage{algorithm}
\usepackage{algpseudocode}
\usepackage{multicol}
\usepackage{caption}

\DeclarePairedDelimiter\autobracket{(}{)}
\newcommand{\br}[1]{\autobracket*{#1}}
\DeclarePairedDelimiter\sparen{[}{]}
\newcommand{\sbr}[1]{\sparen*{#1}}

\newcommand{\R}{\mathbb{R}}
\newcommand{\brt}{\br{\theta}}
\renewcommand{\th}{\theta}
\newcommand{\Var}{\mathbb{V}}
\newcommand{\Cov}{\,\text{Cov}}
\newcommand{\vcov}{\,\text{vcov}}
\renewcommand{\vec}{\text{vec}}
\renewcommand{\ss}{\sigma^2}
\newcommand{\cov}[1]{\,\text{Cov}\br{#1}}

\newcommand{\E}[1]{\mathbb{E}\sbr{#1}}
\newcommand{\Eq}[1]{\mathbb{E}_{#1}}
\DeclareMathOperator*{\argmin}{arg\,min}
\newcommand{\natgrad}{\tilde{\nabla}_\lambda}
\newcommand{\lam}{\lambda}
\newcommand{\equ}{Eq.\,}

\newcommand{\ql}{q_\lambda}
\newcommand{\qlt}{q_\lambda\br{\theta}}
\newcommand{\prior}{p\br{\theta}}
\newcommand{\post}{p\br{\theta\vert y}}

\newcommand{\lik}{p\br{y\vert \theta}}
\renewcommand{\S}{S}
\newcommand{\iS}{S^{-1}}
\newcommand{\q}{q\br{\theta}}
\newcommand{\is}{s^{-1}}

\newcommand{\logq}{\log \q}
\newcommand{\logpy}{\log p\br{y\vert \theta}}

\renewcommand{\th}{\theta}

\newcommand{\iI}{\mathcal{I}^{-1}}
\newcommand{\I}{\mathcal{I}}

\newcommand{\KL}{\text{KL}}
\newcommand{\LB}{\mathcal{L}}
\renewcommand{\vec}{\text{vec}}

\title{Quasi Black-Box Variational Inference with Natural Gradients for Bayesian Learning}

\author{%
  Martin Magris\thanks{Corresponding author. Email: \texttt{magris@ece.au.dk}.}, Mostafa Shabani, Alexandros Iosifidis    \\
  Martin Magris, Mostafa Shabani, Alexandros Iosifidis\\
  Dep. of Electircal and Computing Engineering\\
  Aarhus University\\
  \AA bogade 34, 8200 Aarhus, Denmark \\
}

\begin{document}

\maketitle
\begin{abstract}
We develop an optimization algorithm suitable for Bayesian learning in complex models. Our approach relies on natural gradient updates within a general black-box framework for efficient training with limited model-specific derivations. It applies within the class of exponential-family variational posterior distributions, for which we extensively discuss the Gaussian case for which the updates have a rather simple form. Our Quasi Black-box Variational Inference (QBVI) framework is readily applicable to a wide class of Bayesian inference problems and is of simple implementation as the updates of the variational posterior do not involve gradients with respect to the model parameters, nor the prescription of the Fisher information matrix. We develop QBVI under different hypotheses for the posterior covariance matrix, discuss details about its robust and feasible implementation, and provide a number of real-world applications to demonstrate its effectiveness.
\end{abstract}

\section{Introduction} \label{sec:intro}
Machine Learning (ML) techniques have been proved to be successfully in many prediction and classification tasks across natural-language processing \cite{young_recent_2018}, computer vision \cite{krizhevsky_imagenet_2017}, time-series \cite{langkvist2014review} and finance applications \cite{dixon2020machine}, among the several ones. Recently, Bayesian methods have gained considerable interest in the field as an attractive alternative to point estimation, especially for their ability to address uncertainty via posterior distribution, generalize while reducing overfitting \cite{hoeting1999bayesian}, and for enabling sequential learning \cite{freitas2000hierarchical} while retaining prior and past knowledge. Although Bayesian principles have been proposed in ML decades ago \cite[e.g.][]{mackay1992bayesian,mackay1995probable,lampinen2001bayesian}, it has been only recently that fast and feasible methods boosted a growing use of Bayesian methods in complex models \cite{osawa_practical_2019,khan2018fastscalable,khan2018fastyetsimple}. 

The most challenging task is the computation of the posterior \cite{jospin2022hands}. In the typical ML setting characterized by a high number of parameters and a considerable size of data, traditional sampling methods turn unfeasible, yet approximate methods such as Variational Inference (VI) have been shown to be suitable and successful \cite{saul1996mean,wainwright2008graphical,hoffman2013stochastic,blei2017variational}. Furthermore, recent research advocates the use of the natural gradients for boosting the optimum search and the training \cite{wierstra2014natural}, enabling fast and accurate Bayesian learning algorithms that are scalable and versatile. \\
Stochastic Gradient Descent (SGD) methods \cite{robbins1951stochastic} promoted the use of VI for complex high-dimensional DL models \cite{hoffman2013stochastic,salimans2014using}, yet they require significant implementation and tuning \cite{ranganath2014black}. On the other hand, the use of natural gradients \cite{amari1998natural} within the VI framework under exponential family approximations has been shown to be notably efficient and robust \cite{hoffman2013stochastic,wierstra2014natural,khan2018fastscalable}, often leading to simple updates \cite{khan2018fastyetsimple}. In particular, the common choice of a Gaussian approximation to the true posterior \cite{opper2009variational}, results from \cite{bonnet1964transformations} and \cite{price1958useful} leads to simple updates of the variational mean and precision matrix \cite{khan2017conjugate}. Several algorithms have been derived following these results \cite{khan2018fastscalable}, however, they require the model gradients (and perhaps its hessian), and the positive-definiteness constraint on the posterior covariance matrix is a common problem, see. e.g. \cite{tran2021variational,khan2018fastscalable,lin2020handling,ong2018variational}. \\
Traditional optimization algorithms rely on the extensive use of gradients to dynamically adjust the model weights to minimize a given loss, e.g. through backpropagation. 
This holds also in Bayesian learning based on VI, as the choice of the likelihood and variational approximation determine a cascade of model-specific derivations, not keen to immediately fit a general setting based on a ready-to-use, plug-and-play optimizer. \\
Black-box methods \cite{ranganath2014black} are of straightforward and general use relying on stochastic sampling without model-specific derivations \cite{salimans2014using,paisley2012variational,kingma2013auto}. Even though black-box methods can benefit from numerous improvements such as variance reduction techniques \cite{paisley2012variational,ranganath2014black}, applying natural gradient updates is challenging, as the specification of the Fisher matrix is required.

We propose Quasi Black-box Variational Inference (QBVI) which introduces natural gradients updates within the black-box VI framework. It combines the flexibility of black-box methods with the SGD theory for exponential family VI in a feasible, scalable and flexible optimizer. 
In particular, we rely on VI to approximate the true posterior via parameters' updates that only involve function queries without requiring backpropagation, assumptions on the form of the likelihood, or restrictions on the backbone network. Instead, we employ the typical variational Gaussian assumption \cite{kingma2013auto,khan2018fastyetsimple,osawa_practical_2019}, under which the updates take a rather simple form, embracing both closed-form natural gradient VI elements and black-box elements (thus the word ``quasi'').

We provide results and update rules under both full and diagonal posterior covariance assumptions, and discuss the generalization of the proposed method under the well-established mean-field approximation \cite[e.g.][]{blei2017variational}. We furthermore develop the method of Control variates \cite[e.g.][]{paisley2012variational,salimans2014using} for efficient large-MC sampling, and provide solutions for valid covariance updates adapting developments from existing methods \cite{khan2018fastscalable,tran2021variational,ong2018variational}. Following and discussing the typical practicalities and recommendations in VI applications \cite{tran2021practical}, we provide experiments to validate the proposed optimizer, showing its feasibility in complex learning tasks and promoting its use as a practical and ready-to-use tool for Bayesian optimization.

\section{Variational Inference for Bayesian deep learning}
\subsection{Variational inference}
Let $y$ denote the data and $\lik$ the likelihood of the data based on a postulated model with $\theta \in \Theta$ a $d$-dimensional vector of model parameters. Let $\prior$ be the prior distribution on $\theta$. The goal of Bayesian inference is the posterior distribution $ \post =\prior \lik/p\br{y}$.
Bayesian inference is generally difficult due to the fact that the marginal likelihood  $p\br{y}$ is often intractable and of unknown form. In high-dimensional applications, Monte Carlo (MC) methods for sampling the posterior are challenging and unfeasible, and Variational Inference (VI) is an attractive alternative.

VI consists of an approximate method where the posterior distribution is approximated by a probability density $\q$ (called variational distribution) belonging to some tractable class of distributions $\mathcal{Q}$, such as the exponential family. VI thus turns the Bayesian inference problem into that of finding the best approximation $q^\star \br{\theta} \in \mathcal{Q}$ to $\post$ by minimizing the Kullback-Leibler (KL) divergence from $\q$ to $\post$,
\begin{equation*}
    q^\star = \argmin_{q \in \mathcal{Q}}  \KL\br{q \vert \vert \post} = \argmin_{q \in \mathcal{Q}} \int \q \log \frac{\q}{\post} d\theta \text{.}
\end{equation*}
By simple manipulations, it can be shown the KL minimization problem is equivalent to the maximization problem of the so-called Lower Bound (LB) on $\log p\br{y}$ \cite[e.g.]{tran2021practical},
\begin{equation*}
    \LB\br{q} \coloneqq \int \q \log \frac{\prior \lik}{\q} d \theta 
    = \Eq{q}\sbr{\log \frac{\prior \lik}{\q}} \text{.}
\end{equation*}
For any random vector $\theta$ and a function $g\brt$ we denote by $\Eq{f}[ g\brt]$ the expectation of $g\brt$ where $\theta$ follows a probability distribution with density $f$, i.e. $\Eq{f}[ g\brt] =\Eq{\theta \sim f}[g\brt]$.
To make explicit the dependence of the LB on some vector of parameters $\zeta$ parametrizing the variational posterior we write $\LB\br{\zeta} = \LB\br{q_\zeta} = \Eq{q_\zeta}\sbr{\log p\brt - \log q_\zeta\brt + \logpy  }$.
We operate within the fixed-form variational inference (FFVI) framework, where the parametric form of the variational posterior is set.

\subsection{SGD and natural gradients}
A straightforward approach to maximize the LB is that of using a gradient-based method such as SGD, ADAM, RMSprop. 
In the FFVI setting with $\mathcal{Q}$ being the exponential family, the LB is often optimized in terms of the natural parameter $\lambda$ \cite{wainwright2008graphical}. 
The application of the SGD update based on the standard gradient is problematic because it ignores the information geometry of the distribution $q_\lambda$ \cite{amari1998natural}, as it implicitly relies on the Euclidean norm to capture the dissimilarity between two distributions 
which can indeed be a quite poor and misleading measure of dissimilarity \cite{khan2018fastyetsimple}. By replacing the Euclidean norm with the KL divergence, the SGD update results in the following natural gradient update:
\begin{align}\label{eq:sgd_nat_update}
   \lambda_{t+1} =\lambda_t+\beta_t\left. \sbr{\tilde{\nabla}_\lambda\mathcal{L}\br{\lambda}} \right|_{\lam = \lam_t}
\end{align}
The natural gradient update results in better step directions towards the optimum when optimizing the parameter of a distribution. The natural gradient of $\LB\br{\lambda}$ is obtained by rescaling the euclidean gradient $\nabla_\lambda \LB\br{\lambda}$ by the inverse of the Fisher Information Matrix (FIM) $\I_\lambda$,
\begin{equation} \label{eq:def_natgrad}
  \tilde{\nabla}_\lambda \mathcal{L}\br{\lambda} = \iI_\lambda\nabla_\lam\LB_\lambda \text{.}
\end{equation}
By replacing in the above $\nabla_\lambda \LB\br{\lambda}$ with a stochastic estimate $\hat{\nabla}_\lambda\LB\br{\lambda}$ one obtains a stochastic natural gradient update.

\subsection{Maximization of the Lower Bound}
With respect to an exponential-family prior-posterior pair of the same parametric form, with natural parameters $\eta$ and $\lambda$ respectively, the natural gradient of the LB is given by
\begin{align}
    \natgrad\LB\br{\lambda} &= \natgrad \Eq{\ql}\sbr{\log \frac{p_\eta\brt}{\qlt}} + \natgrad\Eq{\ql}\sbr{\logpy} \label{eq:LB_gradient_generic}\\
    &=  \eta - \lambda +\natgrad\Eq{\ql}\sbr{\logpy}. \label{eq:LB_gradient_khan0}
\end{align}
A proof is provided in Appendix \ref{proof:log_trick}. By applying \eqref{eq:LB_gradient_khan0} to \eqref{eq:sgd_nat_update}, we have the compact update \cite{khan2018fastscalable}
\begin{align}
    \lambda_{t+1} = \br{1-\beta}\lambda_t + \beta\br{\eta + \natgrad\Eq{\ql}\sbr{\logpy}}. \label{eq:update_khan}
\end{align}
Two points are critical. First the $\iI_\lambda$ requirement, which is implicit in the natural gradient definition. Though impractical, it is perhaps possible to estimate $\iI_\lambda$ at each iteration with an iterative conjugate gradient method using only matrix-vector products \cite{tran2021practical}, or exploiting a certain (assumed or imposed) structure of the FIM \cite{tran2020bayesian}. 
Second, the expectation in \eqref{eq:update_khan} is generally intractable as it depends on the specified log-likelihood. Thus, sampling methods can be used.

\subsection{Exponential family basics}
Assume $\qlt$ belongs to an exponential family distribution. Its probability density function is parametrized as
\begin{equation}\label{eq:exp_fam_representation}
    \qlt = h\brt \exp\br{\phi\brt^\top \lambda - A\br{\lambda}}\text{,}
\end{equation}
where $\lambda \in \Omega = \{  \lambda \in \R^d: A\br{\lambda} < +\infty \}$ is the natural parameter, $\phi\brt$ the sufficient statistic, $A\br{\lambda} = \log \int h\brt \exp(\phi\brt^\top \lambda) d\nu$ the log-partition function, determined upon the measure $\nu$, $\phi$, and the function $h$. 
When $\Omega$ is a non-empty open set, the exponential family is referred to as regular. Furthermore, if there are no linear constraints among the components of $\lambda$ and $\phi\brt$, the exponential family \eqref{eq:exp_fam_representation} is of minimal representation. Non-minimal families can always be reduced to minimal families through a suitable transformation and reparametrization, leading to a unique parameter vector $\lambda$ associated with each distribution \cite{wainwright2008graphical}.
The mean (or expectation) parameter $m \in \mathcal{M}$ is defined as a function of $\lambda$, $m\br{\lambda} = \Eq{\ql}\sbr{\phi\brt} = \nabla_\lambda A\br{\lambda}$. Moreover, for the Fisher Information Matrix $\I_\lambda = -\Eq{\ql}\sbr{\nabla_\lambda^2 \log \qlt}$ it holds that $\I_\lambda =\nabla^2_\lambda A\br{\lambda} = \nabla_\lambda m \text{.}$
Under minimal representation, $A\br{\lambda}$ is convex, thus the mapping $\nabla_\lambda A = m:\Omega \rightarrow \mathcal{M}$ is one-to-one, and $\I_\lambda$ is positive definite and invertible \cite{nielsen2009statistical}.
Therefore, under minimal representation we can express $\lambda$ in terms of $m$ and thus $\LB\br{\lambda}$ in terms of $\LB\br{m}$ and vice versa \cite{khan2018fastyetsimple}. Furthermore, for a generic differentiable function $\LB$, by applying the chain rule, $\nabla_\lambda\LB =  \nabla_\lambda m \nabla_m \LB = \nabla_\lam \br{\nabla_\lam A\br{\lam}} \LB= \nabla^2_\lam A\br{\lam} \LB = \mathcal{I}_\lambda \nabla_m \LB$, from which
\begin{equation} 
 \natgrad \LB = \iI_\lambda \nabla_\lambda\LB = \iI_\lam \br{\I_\lam \nabla_m \LB} = \nabla_m \LB \text{,} \label{eq:natgrad_lambda_grad_m}
\end{equation}
expressing the natural gradient in the natural parameter space as the gradient in the expectation parameter space. Eq. \eqref{eq:natgrad_lambda_grad_m} enables the computation of natural gradients in the natural parameter space without requiring the FIM \cite{khan2017conjugate,khan2018fastyetsimple}. 
This translates into the following equivalence:
\begin{align} \label{eq:natgrad2mgrad}
\natgrad\LB = \eta-\lam + \natgrad  \Eq{\ql} \sbr{\logpy} = \eta-\lam + \nabla_m \Eq{\ql}\sbr{\logpy}\text{.}
\end{align}

\section{Proposed method}
\subsection{Quasi-Black-box natural-gradient VI}
Our approach relies on the following equivalence for the natural gradient of the LB: 
\begin{align}
    \natgrad\LB\br{\lambda} &=
     \eta-\lambda +\natgrad\Eq{\ql}\sbr{\logpy} \label{eq:LB_gradient_khan}\\
    &= \eta-\lambda +\Eq{\ql}\sbr{\natgrad\sbr{\log \qlt} \logpy} \label{eq:LB_gradient_martin}\text{.}
\end{align}
A proof is provided in Appendix \ref{proof:log_trick}. The difference between equations \eqref{eq:LB_gradient_khan} and \eqref{eq:LB_gradient_martin} is substantial. Eq. \eqref{eq:LB_gradient_khan} involves the natural gradient with respect to the log-likelihood, thus implicitly the natural gradient with respect to the underlying function of the covariates over which $\lik$ is parametrized. Eq. \eqref{eq:LB_gradient_martin} instead requires the gradients of the score function of the variational distribution only. If $\natgrad \sbr{\logq}$ is available in closed form, \eqref{eq:LB_gradient_martin} prescribes a practical  gradient-free method for updating the posterior's natural parameter. Given that $\natgrad \sbr{\logq}$ has a closed-form, the wording gradient-free is meant in the sense that no further gradients need to be explicitly evaluated, i.e. those of $\logpy$, as it typically happens e.g. in backpropagation: the natural gradient of the log-likelihood w.r.t. the network parameters is not required.
This is the case for several exponential-family distributions.
In section \ref{subsec:nat_grad_free_gaussian_vi} we show that $\natgrad \sbr{\log\qlt}$ is indeed tractable when $\qlt$ is Gaussian. 
For a function $f_\lam$ of the natural parameter, \cite{khan2017conjugate} shows that the natural gradient $\natgrad f_\lam$ corresponds to the gradient $\nabla_mf_m$ where neither the FIM or its inverse are involved (under minimal representation we can express $f_\lam$ in terms of $m$, i.e. $f_m := f_\lam$.  
Eqs. \eqref{eq:LB_gradient_martin} and \eqref{eq:natgrad_lambda_grad_m} therefore lead to the following update:
\begin{align}\label{eq:QBVI_update}
    \lambda_{t+1} &= \br{1-\beta}\lambda_t + \beta\br{\eta + \Eq{\ql}\sbr{\nabla_m \sbr{\log \qlt}\logpy}} \text{.}
\end{align}
We refer to our approach based on the above update as Quasi-Black-box (natural-gradient) Variational Inference (QBVI). Indeed the (natural) gradient of the lower bound involves two ingredients: (i) function queries for $\logpy$ evaluated at some value $\theta$ drawn from the posterior $q\br{\lambda_t}$, without requiring model-specific derivations for $\logpy$ typical in VI applications, and (ii) $\nabla_m \sbr{\log \qlt}$ that needs to be provided. Thus the name quasi-black-box. Algorithm \ref{alg:general_qbvi} summarized the approach in the general case of an exponential-family prior-posterior pair of the same parametric form.

\subsection{QBVI under Gaussian variational posteriors}  \label{subsec:nat_grad_free_gaussian_vi}   

This section focuses on the implementation of the natural gradient update \eqref{eq:QBVI_update} under the typical multivariate Gaussian variational framework, perhaps the most prominent in VI \cite[][]{kucukelbir2015automatic,kingma2013auto,tran2021variational} and Bayesian learning \cite[][]{graves2011practical,blundell2015weight,khan2018fastscalable,osawa_practical_2019}, among the many others.

The multivariate Gaussian  distribution $\mathcal{N}\br{\mu,\S}$ with $d$-dimensional mean vector $\mu$ and covariance matrix $\S$ can be seen as a member of the exponential family \eqref{eq:exp_fam_representation}. Its density reads
\begin{equation*}
    \qlt = \br{2\pi}^{d/2}\exp\{\phi\br{\theta}^\top \lambda - \frac{1}{2}\mu^\top \iS \mu -\frac{1}{2}\log \vert \S\vert\} \text{,}
\end{equation*}
where
\begin{equation*}
    \phi\brt =\begin{bmatrix}
           \theta \\
           \theta\theta^\top
         \end{bmatrix} \text{,}\quad\quad
    \lambda =\begin{bmatrix}
           \lambda_1 \\
           \lambda_2
         \end{bmatrix} =\begin{bmatrix}
           \iS \mu \\
           -\frac{1}{2}\iS
         \end{bmatrix} \text{,}\quad\quad
    m = \begin{bmatrix}
           m_1 \\
           m_2
         \end{bmatrix} = \begin{bmatrix}
          \mu \\
          \S + \mu\mu^\top
         \end{bmatrix} \text{,}                  
\end{equation*}
and $A\br{\lambda} = -\frac{1}{4}\lambda_1^\top\lambda_2^{-1} \lambda_1-\frac{1}{2} \log \br{-2\lambda_2} $. On the other hand, $\zeta = \sbr{\zeta_1^\top, \zeta_2^\top}^\top$ with $\zeta_1 = \mu = m_1$ and $\zeta_2 = S = m_2 - \mu \mu^\top$, constitutes the common parametrization of the multivariate Gaussian distribution in terms of its mean and variance-covariance matrix.

\newtheorem{prop_nat_grad_gaussian}{\textbf{Proposition}}
\begin{prop_nat_grad_gaussian} \label{prop:nat_grad_gaussian}
For a differentiable 
function $\mathcal{L}$, and for $\ql \sim \mathcal{N}\br{\mu,S}$,
\begin{align*}
   \nabla_{m_1} \mathcal{L}  = \nabla_\mu \mathcal{L} -2\sbr{\nabla_S \mathcal{L}}\mu \text{,} \qquad\qquad
   \nabla_{m_2} \mathcal{L}  = \nabla_S \mathcal{L}
\end{align*}
where $m_1 = \mu$ and $m_2 = S + \mu \mu^\top$ are the expectation parameters of $\ql$ \cite{khan2017conjugate}.
\end{prop_nat_grad_gaussian}

By applying the above proposition to \eqref{eq:LB_gradient_khan} and \eqref{eq:LB_gradient_martin}, the computation of the gradient of the LB with respect to $m$,  reduces to evaluating $\nabla_m \logpy$, and thus to its gradients with respect to $\mu$ and $S+\mu \mu^\top$. In the next proposition these gradients are shown to be of a rather simple form.

\begin{prop_nat_grad_gaussian} \label{prop:exp_grad_gaussian_solved}
For $\qlt$ being a Gaussian distribution with mean $\mu$ and covariance matrix $\S$, with natural parameters $\lambda_1$, $\lambda_2$ and expectation parameters $m_1$, $m_2$, be $v = \iS\br{\theta-\mu}$, then
\begin{align*}
    \nabla_{\mu} \log \qlt  = v\text{,} \qquad \qquad     \nabla_{\S} \log \qlt = -\frac{1}{2}\br{\iS - v v^\top}\text{.}
    \end{align*}
    Furthermore,
    \begin{align*}
    \nabla_{m_1} \log \qlt  = \iS \theta - v v^\top\mu \text{,} \qquad \qquad   
    \nabla_{m_2} \log \qlt= \nabla_{\S} \log \qlt \text{.}
\end{align*}
\end{prop_nat_grad_gaussian}
A proof is provided in Appendix \ref{proof:grad_wrt_m}.
Proposition \ref{prop:exp_grad_gaussian_solved} enables the online computation of the updates for the posteriors' natural parameter, in particular 
for a full-covariance Gaussian variational posterior the QBVI  translates in the following updates for $\mu$ and $\iS$: 
\begin{align}
    \iS_{t+1} &= \br{1-\beta}\iS_t +\beta \sbr{\iS_0  + \Eq{\ql}\sbr{\br{\iS_t-v_t v^\top_t}\logpy}} \text{,} \label{eq:update_S_martin} \\
    \mu_{t+1} &= \mu_t + \beta\S_{t+1}[\iS_0\br{\mu_0-\mu_t} + \Eq{\ql}\sbr{v_t\logpy}] \text{,}\label{eq:update_MU_martin}
\end{align}
where $S_0$ and $\mu_0$ respectively denote the mean vector and variance-covariance matrix of the prior distribution. A proof is provided in Appendix \ref{proof:QBVI}. Note that the updates involve a total of $d+d^2$ parameters, and additional $d+d^2$ hyper-parameters are required for the prior specification.
The updates are further simplified under an isotropic Gaussian prior of mean zero $\mu_0=0$ and variance-covariance matrix $S_0 = I/\tau$, with $\tau>0$ a scalar precision parameter. This is a common prior in Bayesian inference, e.g. \cite{ganguly2021introduction,tran2021practical}, and BNN applications, e.g. \cite{graves2011practical,khan2018fastscalable,kingma2013auto}, that furthermore requires the specification of only one prior hyper-parameter.

Furthermore, assuming that the variation posterior is diagonal, the optimization problem reduces to the estimation of $2d$ parameters. Be $s$ and $\is$ the column vector corresponding to the diagonal of $S$ and $\iS$ respectively, the QBVI update reads:
\begin{align}\label{eq:update_S_diagonal}
    \is_{t+1} &= \br{1-\beta}\is_t +\beta \sbr{\bm{\tau}_d  + \Eq{\ql}\sbr{\br{\is_t-v_t \odot  v_t}\logpy}} \\
    \mu_{t+1} &= \mu_t+ \beta s_{t+1}\odot [- \tau \mu_t + \Eq{\ql}\sbr{v_t\logpy}]  \text{,} \nonumber
\end{align}
where $v_t = \is_t \odot\br{\theta-\mu_t}$, $\bm{\tau}_d = (\tau,\dots,\tau)^\top \in \R^d$, and $\odot$ is the element-wise product.
The expectations in \eqref{eq:update_S_martin} and \eqref{eq:update_MU_martin} depend on the chosen likelihood and its parametrization in terms of $\theta$, thus cannot be further simplified. 
While the typical Bayesian framework is here resembled in terms of prior-posterior assumptions, $\logpy$ is unconstrained in its form and complexity, and conjugacy is not a requirement. The updates do not require the gradients of the likelihood, resulting in simple updates (especially under a diagonal posterior and isotropic prior) that exploit natural gradients w.r.t. the variational distribution. Automatic differentiation and backpropagation are here irrelevant potentially enabling the implementation of QBVI in low-level and basic programming languages. The QBVI thus constitutes a generic and ready-to-use solution for Bayesian inference in complex models under a Gaussian variational approximation.
The general QBVI algorithm is summarized in Algorithm \ref{alg:QBVIfull}. The exit function $f_{\text{exit}}$ is discussed in Appendix \ref{app:Z_Appendix_practical_recommendations}.

\begin{figure}
  \begin{minipage}[t]{0.48\textwidth}
    \begin{algorithm}[H]
      \centering
      \caption{General QBVI implementation}\label{alg:general_qbvi}
      \footnotesize
        \begin{algorithmic}[1]
\State \text{Set hyper-parameters: }$ 0<\beta< 1$, $N_s$
\State \text{Set priors and initial values: } $\eta$, $\lambda_0$
\State \text{Set:} $t=1$, $\text{Stop} = \texttt{false}$
\While{$\text{Stop} = \texttt{true}$}
\State \text{Generate: } $\theta_s \sim q_{\lambda_t}$, $s = 1\dots N_s$
\State $\hat{g} = 1/N_s\sum_s \nabla_m \sbr{\log \ql\brt} \logpy$ 
\State $\lambda_{t+1} = \br{1-\beta}\lam_t + \beta\br{\eta + \hat{g}}$
\State $t = t+1$, $\text{Stop} =  f_{\text{exit}}\br{\dots}$ 
\EndWhile
\end{algorithmic}
    \end{algorithm}
  \end{minipage}
  \hfill
  \begin{minipage}[t]{0.48\textwidth}
    \begin{algorithm}[H]
      \centering
      \caption{QBVI, full-covariance Gaussian} \label{alg:QBVIfull} 
      \footnotesize
\begin{algorithmic}[1]
\State \text{Set hyper-parameters: }$ 0<\beta< 1$, $N_s$
\State \text{Set priors and initial values: } $S_0$, $\mu_0$, $S_1$, $\mu_1$
\State \text{Set:} $t=1$, $\text{Stop} = \texttt{false}$
\While{$\text{Stop} = \texttt{true}$}
\State \text{Generate: } $\theta_s \sim q_{\lambda_t}$, $s = 1\dots N_s$
\State $\hat{g}_S = 1/N_s\sum_s \br{\iS_t \theta_s - v_t v_t^\top} \log p\br{y \vert \theta_s}$
\State $\hat{g}_\mu =1/N_s\sum_s v_t \log p\br{y\vert \theta_s}$ 
\State $\iS_{t+1} \leftarrow \br{1-\beta}\iS_t + \beta\sbr{\iS_0 + \hat{g}_S}$, 
\State $\mu_{t+1} \leftarrow \mu_t +\beta S_{t+1}\sbr{\iS_0\br{\mu_0-\mu_t} + \hat{g}_\mu}$ 
\State $t = t+1$, $\text{Stop} =  f_{\text{exit}}\br{\dots}$ 
\EndWhile
\end{algorithmic}
    \end{algorithm}
  \end{minipage}
  \caption*{ \textit{Algorithm \ref{alg:general_qbvi}}. Line 6: estimates the expectation in \eqref{eq:LB_gradient_martin}, line 7: update \eqref{eq:QBVI_update}, line 9: an exit rule, e.g. $f_{\text{exit}}\br{\bar{\LB_t},P,t}$ (see Appendix \ref{app:Z_Appendix_practical_recommendations}). 
  \textit{Algorithm \ref{alg:QBVIfull}}. Lines 6-7: estimate the expectations in \eqref{eq:update_S_martin} and \eqref{eq:update_MU_martin}, lines 8-9: updates \eqref{eq:update_MU_martin} and \eqref{eq:update_S_martin}, line 9: see Algorithm \ref{alg:general_qbvi}.}
\end{figure}

\subsection{Contributions, limitations and related methods}\label{subsec:related_works}

QBVI stands as an algorithm that shares several similarities with several existing alternatives while differing from them as well. In particular, it lies in the gap between feasible natural gradient approaches without requiring the FIM (VON \cite{khan2018fastyetsimple} rationale and exponential family properties) and black-box methods (BBVI\cite{ranganath2014black} rationale and the use of the score estimator) that do not require models' gradients. QBVI provides a solution in this direction. Importantly, the complexity of the underlying model or backbone network is of little relevance for the adoption of QBVI over any suitable likelihood. Whatever the complexity of the underlying backbone model, outputs are parsed to the likelihood. Only forward passes are involved, while QBVI does not constrain the form/complexity of the likelihood. Furthermore, the likelihood does not require differentiability.
In addition, in Appendix \ref{app:proofs} we ease the computation of the optimal control variate coefficient $c^\star_i$ providing an analytic solution for the denominator in \eqref{eq:ci_star} under a Gaussian variational posterior.\\
Among the limitations of our approach we recognize that though practical, the adoption of the score estimator leads to higher variances compared to methods that use the model's gradients. The trade-off is between the efficiency of the estimator and the possibility of feasibility obtaining the model's gradients. As for a large number of VI algorithms, QBVI does not handle the positive-definiteness of the covariance matrix: in Appendix \ref{app:Z_Appendix_Positive_definiteness} we discuss some remedies for the diagonal covariance case.

An extensive overview of our approach compared to related words is provided in Appendix ??. Here we mention that as references for empirically testing QBVI we adopt a Monte Carlo Markov Chain (MCMC) sampler the Black-Bock Variational Inference (BBVI) method of \cite{ranganath2014black}, the Cholesky Gaussian Variational Bayes (CGVB) \cite{titsias2014doubly}, and the Manifold Gaussian Variational Bayes (MGVB) of \cite{tran2021variational}. MCMC is indicative of the true non-variational posterior, BBVI is perhaps the close-most related method, using the log-score trick, but not natural parameters, for Gaussian variational CGVB updates the Cholesky factor with euclidean gradients while adoption of the reparametrization trick (thus model gradients) for controlling the variance of the estimated gradients of the LB. Lastly, MGVB is based on manifold optimization. It uses approximate natural gradients and has the advantage of guaranteeing positive-definiteness of the covariance updated throughout the iterations.

 \label{sec:proposed_method}

\section{Implementation aspects}

\subsection{Gradient estimation and control variates}
For the implementation of the algorithms \ref{alg:general_qbvi} and \ref{alg:QBVIfull} the central aspect to be addressed is the estimation of gradients' expectations and the use of control variates, for which we provide some new analytical results. This is tackled in the following subsection. 
With $\lambda = \lambda_t$ and $\theta_s \sim \ql$, $n = 1,\dots,N_s$, in order to make the update applicable in practice, expectations can be approximated via MC sampling by the following naive estimators:
\begin{align*}
    \Eq{\ql}\sbr{\br{\iS_t - v_t v^\top_t} \logpy }   &\approx \frac{1}{N_s}\sum_{s=1}^{N_s}\sbr{ (\iS_t - \iS_t\br{\theta_s-\mu_t}\br{\theta_s-\mu_t}^\top\iS_t) \log p\br{y \vert \theta_s}}\text{,}\\
    \Eq{\ql}\sbr{v_t \logpy } &\approx \frac{1}{N_s} \sum_{s=1}^{N_s}\sbr{\iS_t \br{\theta_s - \mu_t}\log p\br{y \vert \theta_s}} \text{,}
\end{align*}
where the matrix products are replaced with appropriate element-wise products for the diagonal posterior case.

The variance of the above naive MC estimators can be reduced through the use of Control Variates (CV) \cite{paisley2012variational,salimans2014using}, by approximating $\Eq{\ql}\sbr{\nabla_m\sbr{\log \qlt}\logpy}$ with
$$
\frac{1}{N_s}\sum_{s=1}^{N_s} \nabla_{m_i}\sbr{\log q\br{\theta_s}}  \br{\log p\br{y\vert \theta_s} - c_i} \text{,}
$$
which is an unbiased estimator of the expected gradient but of equal or smaller variance than the naive MC average across samples. For $i=1,2$, the optimal $c_i$ minimizing the variance of the CV estimator is
\begin{equation}\label{eq:ci_star}
    c_i^\star = {\Cov \br{\nabla_{m_i}\sbr{\logq} \logpy,\nabla_{m_i}\logq}}/{\Var\br{\nabla_{m_i}\logq}} \text{.}
\end{equation}
To ensure the unbiasedness of the estimator, the samples used to estimate $c_i^\star$  must be independent of the ones used to estimate the gradient $\nabla_{m_i}\log q\br{\theta_s}$. In practice, this is straightforward, as at iteration $t$, $c_i^\star$ is estimated by reusing samples $\theta_s$ previously drawn from $q_{\lambda_{(t-1)}}$, earlier used to estimate the former CV gradient.
Under a variational Gaussian posterior, the variance terms $\Var\br{\nabla_{m_i}\logq}$ in \eqref{eq:ci_star} are analytically tractable and correspond to
\begin{equation*}
    \Var\br{\nabla_{m_1}\logq}  = \text{diag}\br{\iS\br{S+D}\iS} \text{,} \qquad 
    \Var\br{\nabla_{m_2}\logq} = \frac{1}{4} \vec^{-1}[\text{diag}\br{Q}] \text{.}
\end{equation*}
A proof is provided in Appendix \ref{proof:cv_var}, along with the definition of matrices $D$ and $Q$. The above restricts the MC-based estimation of $c_i^\star$ only to the covariance term in \eqref{eq:ci_star}, clearly intractable as it depends on the general form of $\logpy$. For the CV estimator, we suggest selecting $N_s \approx 100$, as a compromise between estimates' variance and performance. In BNN frameworks exploiting the reparametrization trick \cite{blundell2015weight} a single MC draw can suffice for providing satisfactory estimates of the gradients, but direct comparison is here difficult as it does require the gradient of the loss with respect to each network's parameter, and the expectation is usually taken with respect to a standard normal distribution. Within Appendix \ref{app:Z_Appendix_practical_recommendations} we empirically illustrate the importance of variance control and address the impact of $N_s$ on the variance of the gradients and on the LB.

\subsection{Further considerations for the implementation}
Besides the gradient estimation, there are a number of practicalities and actual implementation details that nevertheless are important that need to be put in place for a smooth implementation. Such standard provisions are collectively discussed in appendix \ref{app:Z_Appendix_practical_recommendations}. Separately in Appendix \ref{app:Z_MFQBVI}, we discuss the implementation of a mean-field variant of QBVI, while in Appendix \ref{app:Z_Appendix_Positive_definiteness} we tackle the issue related to positive definiteness of the variational covariance matrix.

 \label{sec:practical}

\section{Experiments}

The paper introduces a new approach for tacking VI despite the form and complexity of the underlying backbone method. Even for the most elaborate deep-learning applications that may lead to outputs through a sequence of numerous and complex layers, for VB the outputs are parsed into a likelihood function, whose numerical values feed QBVI. In practical applications, the most common forms for the likelihood would those analogous to logistic and linear regression. Indeed a DL network for binary classification would return from its last layer class probabilities for which the very same likelihood of the standard logistic regression applies. Our experiments cover these two common situations and furthermore show that QBVI is applicable (as it should) to non-standard forms of the likelihood function (which is indeed not a part of QBVI, like it is for the class of methods using model's gradients). With this rationale, we decided to suggest experiments that are based on simple models, but relevant (in the form of the likelihood) for more elaborated cases. Extensive results and deeper analyses can be found in Appendix \ref{app:Z_experiments}, along with comparisons between QBVI and the baseline methods introduced in Secection \ref{subsec:related_works}.

We test the QBVI algorithm on different real-world data, in classification and prediction problems. 
The most simple scenario for applying our method is that of the logistic regression. In this case, the updates \eqref{eq:update_MU_martin} and \eqref{eq:update_S_martin} immediately apply, and the likelihood has a simple form. We report our results on two datasets. The Labour datasets \cite{mroz1984sensitivity} consists of 753 individuals and 7 variables, with a binary response variable indicating whether the participants are currently in the labour force.\footnote{Publicly available data:\url{//www.key2stats.com/data-set/view/140}} A 75\%-25\% split is applied to extract the training and testing sets, hyperparameters are provided in Appendix \ref{app:Z_experiments}.

We explore and inspect for anomalies in the dynamics of the learning process for QBVI in Fig. \ref{fig:Labour_LB_mu}.
The smoothed LB ($\bar{\LB}$ of Appendix \ref{app:Z_Appendix_practical_recommendations}) is increasing reaching a plateau in both training and test data after about 1000 epochs (left plot), accordingly the likelihood of the data given the model parameters and the likelihood of the variational posterior display a similar behavior (rightmost plot), suggesting solidity in the learning phase. Standard performance measures display high learning rates at a very early epoch and reach satisfactory steady levels at about epoch 300 already (middle panel). Figure 2 points out a robust and smooth underlying learning process with a relatively fast convergence of the lower bound, variational likelihood, and performance metrics, where the magnitude of MC sampling noise is furthermore relatively small compared to the growth rate of such curves.
Results corresponding to the posterior parameters at the maximum value of the LB are reported in Table 1: 
despite the different optimization objectives (LB as opposed to log-likelihood), the data log-likelihood (LL) under the QBVI and maximum-likelihood (ML) estimates are very close. Performance measures are aligned as well, with rather negligible differences. Note that Table 1 is not meant to address whether the models are adequate and satisfactory in achieving the best performance metrics (LL or others, e.g. accuracy) for the given data and classification task, this is here out of scope. Rather it points out that for the chosen model, QBVI provides a full-Bayesian prescription where the posterior means well-align to the ML estimates leading to negligible differences in the models' LLs evaluated at the ML and QBVI estimates.

\begin{figure}[h!]
    \centering
    \includegraphics[width=0.9\textwidth,trim={4cm 0.5cm 4cm 0.8cm}]{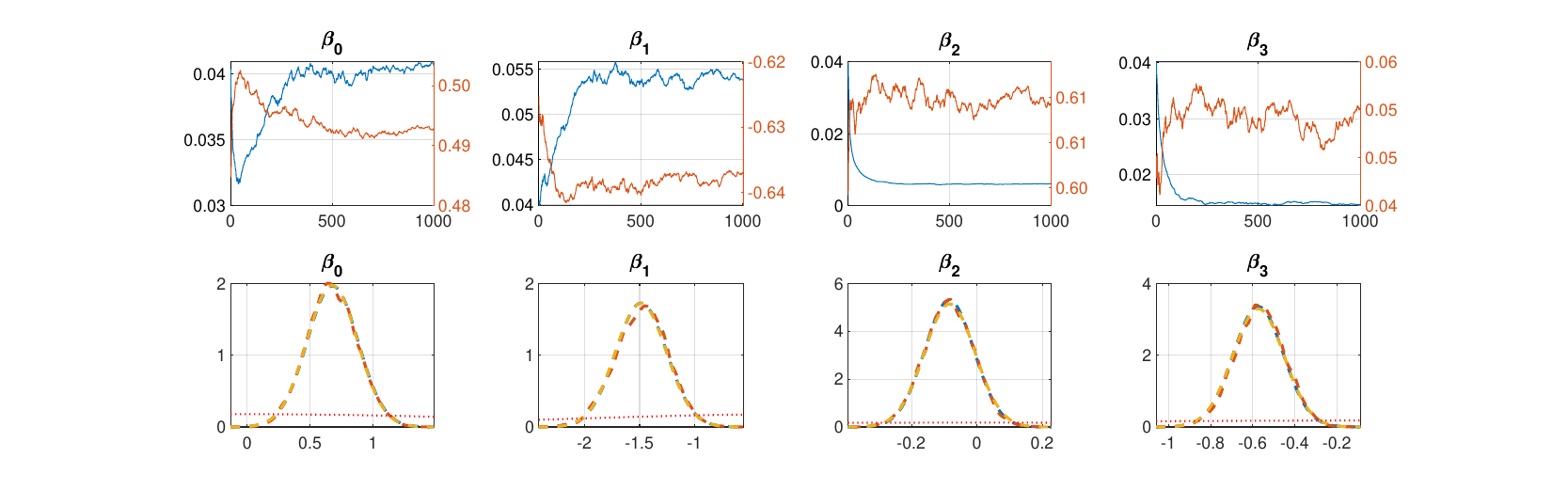}
    \caption{Top panel. Variational posterior means (blue) and variances (orange) of the regression coefficients given in the titles. Bottom row. Variational marginals: QBVI (blue), MCMC samples of the true posterior (orange), MGVI \cite{tran2021variational}. Priors (part of) are overlapped (red).}
    \label{fig:Labour_marginals}
\end{figure}

Figure \ref{fig:Labour_marginals} depicts the learning of the variational parameters and compares the marginal posterior across different models. From the top row, we observe that variational parameters are readily updated within a few iterations and relatively stable later.
For the marginals (bottom row), we observe a remarkable accordance between the QBVI approximation, the Monte-Carlo-Markov-Chain sampler of the true posterior, and the state-of-the-art Gaussian Manifold VI approach \cite{tran2021variational} taken as reference.

The above discussion applies as well to the Statlog (German Credit Data) Data Set\footnote{Publicly available data: \url{archive.ics.uci.edu/ml/datasets/statlog+(german+credit+data)}}, consisting of 24 categorical attributes for 1000 instances for the associated classification task of classifying good or bad credit risk.

Experiments of regression are performed on financial data. We consider the Heterogeneous Autoregressive model (HAR) or realized volatility \cite{corsi2009simple} and the GARCH(1,1) model \cite{bollerslev1986generalized}. For the two, we respectively use daily values of 5-minutes sub-sampled realized volatility \cite{zhang2005tale} and daily close returns for the S\&P 500 index from Jan 1, 2018 to May 15, 2022 (1087 entries).\footnote{Publicly available data:  \url{realized.oxford-man.ox.ac.uk}} The regression results in the bottom part of Table \ref{tab:perf_meas} again show a remarkable alignment between QBVI and maximum likelihood performance measures: it is in the availability of (approximate) joint posterior where the actual difference between two approaches lies. Note that regression requires the estimation of the disturbances' variance, for which we use the mean-field approximation in Algorithm \ref{alg:MFQBVI}, see Appendix \ref{app:Z_MFQBVI}. For all the four datasets, further results are reported in Appendix \ref{app:Z_experiments}.

\begin{figure}
    \centering
    \includegraphics[width=0.6\textwidth,trim={4cm 0.5cm 4cm 1cm}]{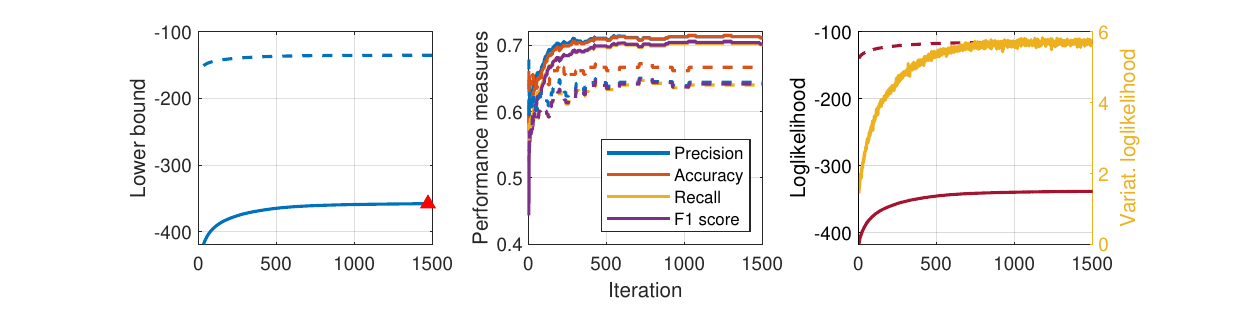}
    \caption{Model learning across iterations. Lower bound (the maximum on the training set marked with a red triangle), performance measures, and loglikelihoods.}
    \label{fig:Labour_LB_mu}
\end{figure}

\begin{table}
    \centering
        \caption{Performance measures for the different datasets. Comparison with the alternative method discussed in Section \ref{subsec:related_works} can be found in Appendix \ref{app:Z_experiments}.}
\scalebox{0.85}{
\begin{tabular}{lrrrrrrrr}
      & \multicolumn{4}{c}{Labour data} & \multicolumn{4}{c}{Credit data} \\
\cmidrule(lr){2-5} \cmidrule(lr){6-9}     & \multicolumn{2}{c}{Train set} & \multicolumn{2}{c}{Test set} & \multicolumn{2}{c}{Train set} & \multicolumn{2}{c}{Test set} \\
   \cmidrule(lr){2-3} \cmidrule(lr){4-5} \cmidrule(lr){6-7}  \cmidrule(lr){8-9} & QBVI  & ML    & QBVI  & ML    & QBVI  & ML    & QBVI  & ML \\
\midrule
Precision & 0.710 & 0.708 & 0.698 & 0.698 & 0.760 & 0.753 & 0.776 & 0.780 \\
Recall & 0.701 & 0.699 & 0.676 & 0.676 & 0.713 & 0.706 & 0.717 & 0.721 \\
Accuracy & 0.711 & 0.709 & 0.612 & 0.612 & 0.791 & 0.785 & 0.656 & 0.667 \\
F1    & 0.703 & 0.701 & 0.746 & 0.746 & 0.728 & 0.721 & 0.815 & 0.816 \\
LL    & -332.99 & -332.98 & -113.89 & -113.80 & -347.68 & -347.48 & -125.80 & -124.50 \\
\midrule
      & \multicolumn{4}{c}{HAR model} & \multicolumn{4}{c}{GARCH model} \\
\cmidrule(lr){2-5} \cmidrule(lr){6-9}       & \multicolumn{2}{c}{Train set} & \multicolumn{2}{c}{Test set} & \multicolumn{2}{c}{Train set} & \multicolumn{2}{c}{Test set} \\
\cmidrule(lr){2-3} \cmidrule(lr){4-5} \cmidrule(lr){6-7}  \cmidrule(lr){8-9} & QBVI  & ML    & QBVI  & ML    & QBVI  & ML    & QBVI  & ML \\
MSE $\times 10^5$ & 10946 & 10945 & 10169 & 10338 & 8.75  & 8.73  & 4.96  & 4.94 \\
LL    & -249.65 & -249.60 & -73.69 & -74.22 & 2587.27 & 2587.34 & 878.02 & 878.04 \\
\bottomrule
\end{tabular}
}
    \label{tab:perf_meas}
\end{table}\label{sec:experiments}

\section{Conclusion}
Whereas Bayesian inference is highly attractive in high-risk domains and applications, several difficulties hinder its use in ML and complex models from a wide audience. Whereas Variational Inference (VI) is an established effective approach for approximating the true posterior with a tractable one, the actual implementation of the parameters' updates is not straightforward. We introduce a Quasi Black-box VI (QBVI) method for the Bayesian learning under a Gaussian variational approximation. Our approach develops on the SGD algorithm with steps in the direction of the natural gradient, to optimize the lower bound on the log-likelihood.

Based on certain properties of the exponential family and of the Gaussian distribution, 
we show that the generally complex natural gradient update can be feasibly approximated by sampling terms that require only function queries of models' likelihood, but not their gradients. Our approach extends the scope of Bayesian learning to the wide class of models for which gradients are difficult or costly to compute, and typical VI model-specific derivations are unfeasible. We provide details on the robust and practical implementation of QBVI update and test its performance on well-established datasets and models. Future research might develop in two directions, virtually extending the current research and overcoming its limitations.
On the other hand, from a theoretical perspective, it is relevant to enable the QBVI update for exploiting the information in structured or factor-decomposed covariance matrices, perhaps resulting in faster updates and increased efficiency, and explore extensions over the Gaussian variational framework. Among the limitations of our work are the trade-off between the convenience of the black-bock approach opposed to alternative making use of models' gradients thus achieving a lower approximation error of the stochastic gradients. Future research direction could explore feasible and effective approaches to variance reduction to be applied in combination, or in alternative, to control variates. In addition the positive-definite constraint is not handled within QBVI and future research could develop in this direction with aid of the manifold optimization theory.\label{sec:conclusion}

\section*{Acknowledgments}
The research received funding from the European Union’s Horizon 2020 research and innovation programme under the Marie Sk{\l}odowska-Curie project BNNmetrics (grant agreement No. 890690).

\newpage
\bibliographystyle{abbrvnat}
\bibliography{Biblio}

\begin{thebibliography}{56}
\providecommand{\natexlab}[1]{#1}
\providecommand{\url}[1]{\texttt{#1}}
\expandafter\ifx\csname urlstyle\endcsname\relax
  \providecommand{\doi}[1]{doi: #1}\else
  \providecommand{\doi}{doi: \begingroup \urlstyle{rm}\Url}\fi

\bibitem[Amari(1998)]{amari1998natural}
S.-I. Amari.
\newblock Natural gradient works efficiently in learning.
\newblock \emph{Neural computation}, 10\penalty0 (2):\penalty0 251--276, 1998.

\bibitem[Anderson and Olkin(1985)]{anderson1985maximum}
T.~W. Anderson and I.~Olkin.
\newblock Maximum-likelihood estimation of the parameters of a multivariate
  normal distribution.
\newblock \emph{Linear algebra and its applications}, 70:\penalty0 147--171,
  1985.

\bibitem[Blei et~al.(2017)Blei, Kucukelbir, and McAuliffe]{blei2017variational}
D.~M. Blei, A.~Kucukelbir, and J.~D. McAuliffe.
\newblock Variational inference: A review for statisticians.
\newblock \emph{Journal of the American statistical Association}, 112\penalty0
  (518):\penalty0 859--877, 2017.

\bibitem[Blundell et~al.(2015)Blundell, Cornebise, Kavukcuoglu, and
  Wierstra]{blundell2015weight}
C.~Blundell, J.~Cornebise, K.~Kavukcuoglu, and D.~Wierstra.
\newblock Weight uncertainty in neural network.
\newblock In \emph{International Conference on Machine Learning}, pages
  1613--1622, 2015.

\bibitem[Bollerslev(1986)]{bollerslev1986generalized}
T.~Bollerslev.
\newblock Generalized autoregressive conditional heteroskedasticity.
\newblock \emph{Journal of econometrics}, 31\penalty0 (3):\penalty0 307--327,
  1986.

\bibitem[Bonnet(1964)]{bonnet1964transformations}
G.~Bonnet.
\newblock Transformations des signaux aléatoires a travers les systèmes non
  linéaires sans mémoire.
\newblock \emph{Annales des Télécommunications}, 19\penalty0 (9-10):\penalty0
  203--220, 1964.

\bibitem[{\c{C}}inlar(2011)]{ccinlar2011probability}
E.~{\c{C}}inlar.
\newblock \emph{Probability and stochastics}, volume 261.
\newblock Springer, 2011.

\bibitem[Corsi(2009)]{corsi2009simple}
F.~Corsi.
\newblock A simple approximate long-memory model of realized volatility.
\newblock \emph{Journal of Financial Econometrics}, 7\penalty0 (2):\penalty0
  174--196, 2009.

\bibitem[Dixon et~al.(2020)Dixon, Halperin, and Bilokon]{dixon2020machine}
M.~F. Dixon, I.~Halperin, and P.~Bilokon.
\newblock \emph{Machine learning in Finance}, volume 1170.
\newblock Springer, 2020.

\bibitem[Freitas et~al.(2000)Freitas, Niranjan, and
  Gee]{freitas2000hierarchical}
J.~F. G.~d. Freitas, M.~Niranjan, and A.~H. Gee.
\newblock Hierarchical bayesian models for regularization in sequential
  learning.
\newblock \emph{Neural computation}, 12\penalty0 (4):\penalty0 933--953, 2000.

\bibitem[Ganguly and Earp(2021)]{ganguly2021introduction}
A.~Ganguly and S.~W.~F. Earp.
\newblock An introduction to variational inference.
\newblock \emph{arXiv preprint arXiv:2108.13083}, 2021.

\bibitem[Ghosh and Sinha(2002)]{ghosh2002simple}
M.~Ghosh and B.~K. Sinha.
\newblock A simple derivation of the wishart distribution.
\newblock \emph{The American Statistician}, 56\penalty0 (2):\penalty0 100--101,
  2002.

\bibitem[Graves(2011)]{graves2011practical}
A.~Graves.
\newblock Practical variational inference for neural networks.
\newblock \emph{Advances in neural information processing systems}, 24, 2011.

\bibitem[Hoeting et~al.(1999)Hoeting, Madigan, Raftery, and
  Volinsky]{hoeting1999bayesian}
J.~A. Hoeting, D.~Madigan, A.~E. Raftery, and C.~T. Volinsky.
\newblock Bayesian model averaging: a tutorial.
\newblock \emph{Statistical science}, 14\penalty0 (4):\penalty0 382--417, 1999.

\bibitem[Hoffman et~al.(2013)Hoffman, Blei, Wang, and
  Paisley]{hoffman2013stochastic}
M.~D. Hoffman, D.~M. Blei, C.~Wang, and J.~Paisley.
\newblock Stochastic variational inference.
\newblock \emph{Journal of Machine Learning Research}, 2013.

\bibitem[Jospin et~al.(2022)Jospin, Laga, Boussaid, Buntine, and
  Bennamoun]{jospin2022hands}
L.~V. Jospin, H.~Laga, F.~Boussaid, W.~Buntine, and M.~Bennamoun.
\newblock Hands-on bayesian neural networks — a tutorial for deep learning
  users.
\newblock \emph{IEEE Computational Intelligence Magazine}, 17\penalty0
  (2):\penalty0 29--48, 2022.

\bibitem[Khan and Lin(2017)]{khan2017conjugate}
M.~Khan and W.~Lin.
\newblock Conjugate-computation variational inference: Converting variational
  inference in non-conjugate models to inferences in conjugate models.
\newblock In \emph{Artificial Intelligence and Statistics}, pages 878--887,
  2017.

\bibitem[Khan et~al.(2018)Khan, Nielsen, Tangkaratt, Lin, Gal, and
  Srivastava]{khan2018fastscalable}
M.~Khan, D.~Nielsen, V.~Tangkaratt, W.~Lin, Y.~Gal, and A.~Srivastava.
\newblock Fast and scalable bayesian deep learning by weight-perturbation in
  adam.
\newblock In \emph{International Conference on Machine Learning}, pages
  2611--2620, 2018.

\bibitem[Khan and Nielsen(2018)]{khan2018fastyetsimple}
M.~E. Khan and D.~Nielsen.
\newblock Fast yet simple natural-gradient descent for variational inference in
  complex models.
\newblock In \emph{2018 International Symposium on Information Theory and Its
  Applications}, pages 31--35, 2018.

\bibitem[Khan and Rue(2020)]{khan2020learning}
M.~E. Khan and H.~Rue.
\newblock Learning algorithms from bayesian principles.
\newblock \emph{Draft v. 0.7, August}, 2020.

\bibitem[Khan and Rue(2021)]{khan2021bayesian}
M.~E. Khan and H.~Rue.
\newblock The bayesian learning rule.
\newblock \emph{arXiv preprint arXiv:2107.04562}, 2021.

\bibitem[Kingma and Welling(2014)]{kingma2013auto}
D.~P. Kingma and M.~Welling.
\newblock Auto-encoding variational bayes.
\newblock \emph{International Conference on Learning Representations}, 2014.

\bibitem[Krizhevsky et~al.(2012)Krizhevsky, Sutskever, and
  Hinton]{krizhevsky_imagenet_2017}
A.~Krizhevsky, I.~Sutskever, and G.~E. Hinton.
\newblock Imagenet classification with deep convolutional neural networks.
\newblock \emph{Advances in neural information processing systems}, 25, 2012.

\bibitem[Kucukelbir et~al.(2015)Kucukelbir, Ranganath, Gelman, and
  Blei]{kucukelbir2015automatic}
A.~Kucukelbir, R.~Ranganath, A.~Gelman, and D.~Blei.
\newblock Automatic variational inference in stan.
\newblock \emph{Advances in neural information processing systems}, 28, 2015.

\bibitem[Lampinen and Vehtari(2001)]{lampinen2001bayesian}
J.~Lampinen and A.~Vehtari.
\newblock Bayesian approach for neural networks—review and case studies.
\newblock \emph{Neural networks}, 14\penalty0 (3):\penalty0 257--274, 2001.

\bibitem[L{\"a}ngkvist et~al.(2014)L{\"a}ngkvist, Karlsson, and
  Loutfi]{langkvist2014review}
M.~L{\"a}ngkvist, L.~Karlsson, and A.~Loutfi.
\newblock A review of unsupervised feature learning and deep learning for
  time-series modeling.
\newblock \emph{Pattern Recognition Letters}, 42:\penalty0 11--24, 2014.

\bibitem[Lehmann and Casella(2006)]{lehmann2006theory}
E.~L. Lehmann and G.~Casella.
\newblock \emph{Theory of point estimation}.
\newblock Springer Science \& Business Media, 2006.

\bibitem[Lin et~al.(2020)Lin, Schmidt, and Khan]{lin2020handling}
W.~Lin, M.~Schmidt, and M.~E. Khan.
\newblock Handling the positive-definite constraint in the bayesian learning
  rule.
\newblock In \emph{International Conference on Machine Learning}, pages
  6116--6126, 2020.

\bibitem[Lyu and Tsang(2021)]{lyu2021black}
Y.~Lyu and I.~W. Tsang.
\newblock Black-box optimizer with stochastic implicit natural gradient.
\newblock In \emph{Joint European Conference on Machine Learning and Knowledge
  Discovery in Databases}, pages 217--232, 2021.

\bibitem[Mackay(1992)]{mackay1992bayesian}
D.~J.~C. Mackay.
\newblock \emph{Bayesian methods for adaptive models}.
\newblock PhD thesis, California Institute of Technology, 1992.

\bibitem[Mackay(1995)]{mackay1995probable}
D.~J.~C. Mackay.
\newblock Probable networks and plausible predictions — a review of practical
  bayesian methods for supervised neural networks.
\newblock \emph{Network: Computation In Neural Systems}, 6:\penalty0 469--505,
  1995.

\bibitem[Mohamed et~al.(2020)Mohamed, Rosca, Figurnov, and
  Mnih]{mohamed2020monte}
S.~Mohamed, M.~Rosca, M.~Figurnov, and A.~Mnih.
\newblock Monte carlo gradient estimation in machine learning.
\newblock \emph{Journal of Machine Learning Research}, 21\penalty0
  (132):\penalty0 1--62, 2020.

\bibitem[Mroz(1984)]{mroz1984sensitivity}
T.~A. Mroz.
\newblock \emph{The sensitivity of an empirical model of married women's hours
  of work to economic and statistical assumptions}.
\newblock PhD thesis, Stanford University, 1984.

\bibitem[Muirhead(2009)]{muirhead2009aspects}
R.~J. Muirhead.
\newblock \emph{Aspects of multivariate statistical theory}.
\newblock John Wiley \& Sons, 2009.

\bibitem[Nielsen and Garcia(2009)]{nielsen2009statistical}
F.~Nielsen and V.~Garcia.
\newblock Statistical exponential families: A digest with flash cards.
\newblock \emph{arXiv preprint arXiv:0911.4863}, 2009.

\bibitem[Ong et~al.(2018)Ong, Nott, Tran, Sisson, and
  Drovandi]{ong2018variational}
V.~M.~H. Ong, D.~J. Nott, M.-N. Tran, S.~A. Sisson, and C.~C. Drovandi.
\newblock Variational bayes with synthetic likelihood.
\newblock \emph{Statistics and Computing}, 28:\penalty0 971--988, 2018.

\bibitem[Opper and Archambeau(2009)]{opper2009variational}
M.~Opper and C.~Archambeau.
\newblock The variational gaussian approximation revisited.
\newblock \emph{Neural computation}, 21\penalty0 (3):\penalty0 786--792, 2009.

\bibitem[Osawa et~al.(2019)Osawa, Swaroop, Khan, Jain, Eschenhagen, Turner, and
  Yokota]{osawa_practical_2019}
K.~Osawa, S.~Swaroop, M.~E. Khan, A.~Jain, R.~Eschenhagen, R.~E. Turner, and
  R.~Yokota.
\newblock Practical deep learning with bayesian principles.
\newblock In \emph{Advances in Neural Information Processing Systems},
  volume~32, pages 1--13, 2019.

\bibitem[Paisley et~al.(2012)Paisley, Blei, and Jordan]{paisley2012variational}
J.~Paisley, D.~M. Blei, and M.~I. Jordan.
\newblock Variational bayesian inference with stochastic search.
\newblock \emph{arXiv preprint arXiv:1206.6430}, 2012.

\bibitem[Price(1958)]{price1958useful}
R.~Price.
\newblock A useful theorem for nonlinear devices having gaussian inputs.
\newblock \emph{IRE Transactions on Information Theory}, 4\penalty0
  (2):\penalty0 69--72, 1958.

\bibitem[Ranganath et~al.(2014)Ranganath, Gerrish, and
  Blei]{ranganath2014black}
R.~Ranganath, S.~Gerrish, and D.~M. Blei.
\newblock Black box variational inference.
\newblock In \emph{Artificial intelligence and statistics}, pages 814--822,
  2014.

\bibitem[Robbins and Monro(1951)]{robbins1951stochastic}
H.~Robbins and S.~Monro.
\newblock A stochastic approximation method.
\newblock \emph{The annals of mathematical statistics}, pages 400--407, 1951.

\bibitem[Salimans and Knowles(2014)]{salimans2014using}
T.~Salimans and D.~A. Knowles.
\newblock On using control variates with stochastic approximation for
  variational bayes and its connection to stochastic linear regression.
\newblock \emph{arXiv preprint arXiv:1401.1022}, 2014.

\bibitem[Saul et~al.(1996)Saul, Jaakkola, and Jordan]{saul1996mean}
L.~K. Saul, T.~Jaakkola, and M.~I. Jordan.
\newblock Mean field theory for sigmoid belief networks.
\newblock \emph{Journal of artificial intelligence research}, 4:\penalty0
  61--76, 1996.

\bibitem[Tan(2021)]{tan2021natural}
L.~S. Tan.
\newblock Natural gradient updates for cholesky factor in gaussian and
  structured variational inference.
\newblock \emph{arXiv preprint arXiv:2109.00375}, 2021.

\bibitem[Tan and Nott(2018)]{tan2018gaussian}
L.~S. Tan and D.~J. Nott.
\newblock Gaussian variational approximation with sparse precision matrices.
\newblock \emph{Statistics and Computing}, 28\penalty0 (2):\penalty0 259--275,
  2018.

\bibitem[Titsias and L{\'a}zaro-Gredilla(2014)]{titsias2014doubly}
M.~Titsias and M.~L{\'a}zaro-Gredilla.
\newblock Doubly stochastic variational bayes for non-conjugate inference.
\newblock In \emph{International conference on machine learning}, pages
  1971--1979. PMLR, 2014.

\bibitem[Tran et~al.(2019{\natexlab{a}})Tran, Iosifidis, Kanniainen, and
  Gabbouj]{tran_temporal_2019}
D.~T. Tran, A.~Iosifidis, J.~Kanniainen, and M.~Gabbouj.
\newblock Temporal {Attention}-{Augmented} {Bilinear} {Network} for {Financial}
  {Time}-{Series} {Data} {Analysis}.
\newblock \emph{IEEE Transactions on Neural Networks and Learning Systems},
  30\penalty0 (5):\penalty0 1407--1418, 2019{\natexlab{a}}.

\bibitem[Tran et~al.(2019{\natexlab{b}})Tran, Nguyen, Nott, and
  Kohn]{tran2020bayesian}
M.-N. Tran, N.~Nguyen, D.~J. Nott, and R.~Kohn.
\newblock Bayesian deep net glm and glmm.
\newblock \emph{Journal of Computational and Graphical Statistics},
  29:\penalty0 113--97, 2019{\natexlab{b}}.

\bibitem[Tran et~al.(2021{\natexlab{a}})Tran, Nguyen, and
  Nguyen]{tran2021variational}
M.-N. Tran, D.~H. Nguyen, and D.~Nguyen.
\newblock Variational bayes on manifolds.
\newblock \emph{Statistics and Computing}, 31\penalty0 (6):\penalty0 1--17,
  2021{\natexlab{a}}.

\bibitem[Tran et~al.(2021{\natexlab{b}})Tran, Nguyen, and
  Dao]{tran2021practical}
M.-N. Tran, T.-N. Nguyen, and V.-H. Dao.
\newblock A practical tutorial on variational bayes.
\newblock \emph{arXiv preprint arXiv:2103.01327}, 2021{\natexlab{b}}.

\bibitem[Trusheim et~al.(2018)Trusheim, Condurache, and
  Mertins]{trusheim2018boosting}
F.~Trusheim, A.~Condurache, and A.~Mertins.
\newblock Boosting black-box variational inference by incorporating the natural
  gradient.
\newblock In \emph{2018 24th International Conference on Pattern Recognition
  (ICPR)}, pages 19--24. IEEE, 2018.

\bibitem[Wainwright and Jordan(2008)]{wainwright2008graphical}
M.~J. Wainwright and M.~I. Jordan.
\newblock Graphical models, exponential families, and variational inference.
\newblock \emph{Foundations and Trends{\textregistered} in Machine Learning},
  1\penalty0 (1--2):\penalty0 1--305, 2008.

\bibitem[Wierstra et~al.(2014)Wierstra, Schaul, Glasmachers, Sun, Peters, and
  Schmidhuber]{wierstra2014natural}
D.~Wierstra, T.~Schaul, T.~Glasmachers, Y.~Sun, J.~Peters, and J.~Schmidhuber.
\newblock Natural evolution strategies.
\newblock \emph{The Journal of Machine Learning Research}, 15\penalty0
  (1):\penalty0 949--980, 2014.

\bibitem[Young et~al.(2018)Young, Hazarika, Poria, and
  Cambria]{young_recent_2018}
T.~Young, D.~Hazarika, S.~Poria, and E.~Cambria.
\newblock Recent trends in deep learning based natural language processing.
\newblock \emph{IEEE Computational Intelligence Magazine}, 13\penalty0
  (3):\penalty0 55--75, 2018.

\bibitem[Zhang et~al.(2005)Zhang, Mykland, and A{\"\i}t-Sahalia]{zhang2005tale}
L.~Zhang, P.~A. Mykland, and Y.~A{\"\i}t-Sahalia.
\newblock A tale of two time scales: Determining integrated volatility with
  noisy high-frequency data.
\newblock \emph{Journal of the American Statistical Association}, 100\penalty0
  (472):\penalty0 1394--1411, 2005.

\end{thebibliography}

\newpage
\section*{Checklist}


\begin{enumerate}

\item For all authors...
\begin{enumerate}
  \item Do the main claims made in the abstract and introduction accurately reflect the paper's contributions and scope?
    \answerYes{}
  \item Did you describe the limitations of your work?
    \answerYes{Specifically across the text we mention: (i) Actual difficulty in dealing with full-covariance matrices in large-scale problems, leading to diagonal updates (see Sec. \ref{sec:proposed_method} and Sec. \eqref{eq:update_S_diagonal}), (ii) Difficulty in guaranteeing positive-definiteness in covariance updates (Appendix \ref{app:Z_Appendix_Positive_definiteness}), (iii) The simple natural gradient computation is applicable to Gaussian variational approximations only (see \ref{sec:proposed_method} and \ref{sec:conclusion}). The variance of the log-score estimator of the expected gradient may be higher compared to e.g. the reparametrization trick where models' gradients are used}.
  \item Did you discuss any potential negative societal impacts of your work?
    \answerNA{}
  \item Have you read the ethics review guidelines and ensured that your paper conforms to them?
    \answerYes{The ethic guidelines are not a concern for the proposed research and the form it is here presented.}
\end{enumerate}

\item If you are including theoretical results...
\begin{enumerate}
  \item Did you state the full set of assumptions of all theoretical results?
    \answerYes{Proofs in Appendix \ref{app:proofs}}.
        \item Did you include complete proofs of all theoretical results?
    \answerYes{}
\end{enumerate}

\item If you ran experiments...
\begin{enumerate}
  \item Did you include the code, data, and instructions needed to reproduce the main experimental results (either in the supplemental material or as a URL)?
    \answerYes{Codes included in the submission as additional material, and planned to be made available online in the future.}
  \item Did you specify all the training details (e.g., data splits, hyperparameters, how they were chosen)?
    \answerYes{See Sec.\ref{sec:experiments} and Appendix. \ref{app:Z_experiments}}
        \item Did you report error bars (e.g., with respect to the random seed after running experiments multiple times)?
    \answerYes{Rather than error bars we deal with distributions, and report details e.g. on the marginals.}
        \item Did you include the total amount of compute and the type of resources used (e.g., type of GPUs, internal cluster, or cloud provider)?
    \answerYes{We report the run-times in Table \ref{tab:par_credit}, for the largest model. Time statistics are deemed to be taken as indicative as MGVB is efficiently implemented in the BVayes-lab package, wheres out current implementation focuses back-testing. We are currently developing an efficient solution for GPU computations in Python. Codes will be made publicly available.}
\end{enumerate}

\item If you are using existing assets (e.g., code, data, models) or curating/releasing new assets...
\begin{enumerate}
  \item If your work uses existing assets, did you cite the creators?
    \answerYes{}
  \item Did you mention the license of the assets?
    \answerYes{}
  \item Did you include any new assets either in the supplemental material or as a URL?
    \answerYes{}
  \item Did you discuss whether and how consent was obtained from people whose data you're using/curating?
    \answerYes{}
  \item Did you discuss whether the data you are using/curating contains personally identifiable information or offensive content?
    \answerNA{}
\end{enumerate}

\item If you used crowdsourcing or conducted research with human subjects...
\begin{enumerate}
  \item Did you include the full text of instructions given to participants and screenshots, if applicable?
    \answerNA{}
  \item Did you describe any potential participant risks, with links to Institutional Review Board (IRB) approvals, if applicable?
    \answerNA{}
  \item Did you include the estimated hourly wage paid to participants and the total amount spent on participant compensation?
    \answerNA{}
\end{enumerate}

\end{enumerate}

\newpage
\appendix

\section{QBVI in the context of the related literature}

With respect to the NGVI update of \cite{khan2018fastscalable} QBVI differs by the use of the log-score estimator \equ\eqref{eq:LB_gradient_martin} does not require the differentiability of the log-likelihood. 
At any point in the learning process, even though with zero probability, the current value of $\theta_t$ might be such that the gradient $\iI \nabla_\lambda \Eq{q_\lam}\sbr{\logpy}$ in \equ\eqref{eq:LB_gradient_khan} does not exist and NGVI unfeasible. This e.g. would apply to any DL architecture using e.g. ReLU or binary-step activation functions. 
In practice, this would be irrelevant as in a setting where the expectations are estimated $N$ MC samples would occur only if the sample values are the same and equal to the point $\theta$ where the function is non-differentiable.
The difference with respect NGVI derived method such as VON, VOGN, VADAM and AdaGrad \cite{khan2018fastscalable,khan2018fastyetsimple,osawa_practical_2019} methods is the simplicity of QBVI where the use of the log-score estimator avoids the computation of model's gradient ($g$) and hessian ($H$) of the (negative) log-likelihood (and the corresponding assumptions on the log-likelihood) for evaluating $\nabla_\mu \LB$ and $\nabla_\S \LB$ respectively as $\Eq{q}\sbr{g\brt}$ and $\frac{1}{2} \Eq{q}\sbr{H\brt}$ \cite{opper2009variational}, arising when applying \equ\ref{prop:nat_grad_gaussian} to \equ\eqref{eq:natgrad_lambda_grad_m} \cite[appendices E and D]{khan2018fastscalable}. 

With respect to the CVI method of \cite{khan2017conjugate}, QBVI uses the log-score gradient estimator while CIV requires the computation of the model's gradients and theoretical assumption on the function $\LB$ and its gradients (detail in \cite{khan2017conjugate}). Furthermore, CVI uses stochastic approximation of the FIM and gradients of the log-likelihood, while QVI relies on Eq.\ref{eq:natgrad_lambda_grad_m} to handle the exact computation of the natural gradient without requiring the FIM. With respect to the CVI for mean-field approximations, QBVI does not make a distinction between conjugate and non-conjugate terms: in QBVI the prior-posterior pair is assumed of the same form generally resulting in non-conjugate computations. If the prior-(true) posterior is conjugate, adopting a variational approximating for the posterior would generally turn the inference non-conjugate, unless the chosen form of the variational posterior is the same as the true posterior, in which case the variational approximation would be useless and the inference problem turn perhaps analytically tractable. 
Over the MGVB approach of \cite{tran2021variational}, QBVI uses exact natural gradients (w.r.t. $\br{\lam_1,\lam_2}$), whereas MGVB an approximate version (w.r.t. $\br{\mu,\S}$). MGVB is however granted to provide a positive definite update of $\iS$ while QBVI, as the methods described above, does not. Positive-definite issues are tackled in \cite{lin2020handling}, where the Bayesian learning rule\cite{khan2020learning} is augmented with an additional term meant to grant the constraint on $\lam$. 
Despite QBVI model gradients, thus model-specific derivations are still involved both in \cite{lin2020handling} and \cite{khan2020learning}. 
Also, methods updating the Cholesky factor do require model gradients, as based on the reparametrization trick \cite{titsias2014doubly,tan2018gaussian,tan2021natural}

Applying the score-estimator within the algorithm of \cite{lin2020handling}
is certainly a direction to explore for rendering QBVI robust to positive-definite issues in the update of $\lam_2$. 
Without referring to the exponential family, the use of SGD is discussed as well in \cite{wierstra2014natural} with the use of a score-function estimator, 
yet the FIM is approximated with MC sampling (and its inversion is nevertheless required)
here inefficient in light of \equ\eqref{eq:natgrad_lambda_grad_m}. 
\cite{lyu2021black} uses an NGVI update for tacking the general relaxed optimization problem of minimizing an expectation $\Eq{q}\sbr{f\brt}$ with their INGO algorithm. QBVI can be seen as a specific adaptation for the purpose of Bayesian inference. With $f$ being a generic function, in \cite{lyu2021black} it is not decomposed into $\prior\logpy$, leads to the VI-ready updates \eqref{eq:update_S_martin}, \eqref{eq:update_MU_martin}.
Our paper furthermore adopts variance control and tackles the positive-definiteness constraint for the diagonal covariance case. For VI, QBVI is preferred  as the variance of the MC gradients is reduced: INGO samples \equ\eqref{eq:LB_gradient_khan0}, QBVI \equ\eqref{eq:LB_gradient_generic}.
\cite{trusheim2018boosting} also provides a closely related algorithm, based on an approximate MC estimation of the FIM (later inverted). On the other hand, we use the results on exponential families to achieve an exact natural gradient update that implicitly avoids the computation of the inverse FIM. Both QBVI and \cite{trusheim2018boosting} use control variates 
yet we provide an analytic treatment of the denominator involved in computing the optimal coefficients $c_i^\star$. 

QBVI closely relates to BBVI \cite{ranganath2014black}. QBVI introduces is the use of natural gradients, that in the context of exponential-family distributions do not require the computation of $\iI$.
Also, the simplification argument in Appendix \ref{app:proofs} of the LB gradient under the same parametric form of the prior-variational posterior pair immediately adapts to euclidean gradients improving the efficiency of BBVI. Note that there are considerable issues in adopting euclidean gradients for Gaussian VI, especially related to the initialization of the variational posterior and the impossibility to precisely locate quadratic-form optima \cite{wierstra2014natural}.
A direct comparison with SVI \cite{hoffman2013stochastic}, is challenging as SVI applies to a specific class of graphical model where the distinction between local and global hidden variables and natural parameters at a variational distribution level turn the natural gradient computation tractable but qualitatively different form QBVI. 
Furthermore, in the topic modeling context of \cite{hoffman2013stochastic} it is rather inappropriate to assume the same parametric form for the prior and variational posterior. 

As noted in \cite{lyu2021black}, there is a further subtle difference w.r.t. MGVB, BBVI, and SVI, the methods relying on the NGVI update (e.g. QBVI, VON, INGO). In the latter ones the natural gradients update the natural parameters,
Indeed the NGVI updates are provided for $\br{\mu,\Sigma}$ as $\br{\lam_1, \lam_2}$ is not a sufficient prescription for generating samples according to the variational posterior.
In particular, for the class of NGVI-related models (including QBVI) $\mu$ is retrieved from the definition of $\lam_1$ from the updated $\lam_1$.

Among the works that do consider variance control, control variates are predominant, see e.g. \cite{trusheim2018boosting,ranganath2014black,tran2021variational,tran2020bayesian} and \cite[Section 3.7]{tran2021practical}. The use of the reparametrization trick \cite{kingma2013auto} falls outside the scope of providing a general algorithm as model-specific gradients are required. \cite{ranganath2014black} discusses partial averaging by exploiting the mean-field structure of their variational posterior. In a non-mean-field setting partial averaging appears inapplicable, though it could be partially exploited in BNN with a layer-wise block-diagonal posterior covariance matrix ignoring variables' correlations between different layers (for a Gaussian variational posterior this is indeed a mean-field specification where layers are treated independently \cite[Appendix B.3]{osawa_practical_2019}). 
We adopt control variates as they are predominant in the relevant literature and their implementation is immediate, a perhaps feasible alternative could be retrieved by adapting the doubly stochastic variational Bayes principle of \cite{titsias2014doubly}. Alternative strategies for variance control, can be found e.g. in \cite{mohamed2020monte}.\\

\section{Implementation aspects and practical recommendations} \label{app:Z_Appendix_practical_recommendations}

\textit{Classification, regression, and mean-field QBVI}. In general classification problems models' outputs are based on the class of predicted maximum probability, and implicitly rely on a softmax activation function at the last layer. In this setting, the QBVI update outlined above immediately applies as $\logpy$ corresponds to the log-likelihood loss $\sum_{i=i}^M y_{i,c} \log p_i\br{c}$, with $y_{i,c}$ representing a one-key-hot encoding of the $i$-th target with true class $c$, and $p_i\br{c}$ the predicted probability of class $c$ for the $i$-th sample. Though the parametric form of $\logpy$ is different, for certain regression problems such as Binomial or Poisson regression, the outlined QBVI update applies as well. 
 However, additional parameters might enter into play. For instance, take the linear regression: we can explicitly write the Gaussian likelihood as $\log p\br{y|f_\theta\br{x},v}$, and the variational posterior as $q\br{\lambda,\nu}$, with $f_\theta$ represents the underlying regression with weights $\theta$.
 The residuals' variance $v$ is generally unknown and needs to be estimated. In Appendix \ref{app:Z_MFQBVI} we show that by introducing a corresponding additional variational parameter, the Bayesian inference on the posterior $p\br{\theta_1,v\vert y}p\br{v\vert y}$ is readily enabled within a straightforward mean-field form of the QBVI update.
 
 \textit{Constraints on model parameters}. Imposing constraints on the back-bone network parameter is straightforward and does not require modifications for the QBVI update.
 In QBVI, network weights are sampled from the variational Gaussian, defined on the real line. Assume that a constraint is imposed on a network weight $w$ such that it should lie on a support $\mathcal{S}$. It suffices to identify a suitable transform $f:\R \rightarrow \mathcal{S}$ and feed-forward the network by applying $w=f\br{\theta}$. 
 Of course, the implication of applying transformations is that the Gaussian variational assumption holds for $\theta = f^{-1}\br{w}$ (e.g. logit or log of $w$ if $f$ is the sigmoid or exponential function, respectively aiming at guaranteeing $0<w<1$ and $w>0$), rather than the actual model parameter $w$ for which the variational approximation is $\mathcal{N}\br{f^{-1}\br{\theta}; \mu,S}\vert \text{det}\br{J_{f^{-1}}\br{\theta}}\vert$, with $J_{f^{-1}}$ the Jacobin of the inverse transform \cite{kucukelbir2015automatic}.
 \textit{Example.} For the mixing coefficient $w$ the TABL layer \cite{tran_temporal_2019}, the constraint $0<w<1$ applies. The sigmoid function maps $\R$ to $\br{0,1}$: by replacing $w$ with $f\br{\theta} = 1/(1+\exp\br{-\theta})$, although $\theta \in \R$, $0<w<1$ holds. \textit{Example.} A more elaborate example is described with Table \ref{tab:par_GARCH}.
 
 \textit{LB smoothing and stopping criterion}. At each iteration the LB is expected to improve, yet the stochastic nature of the estimates $\hat{\LB}\br{\lambda}$ introduces noise that can violate its expected non-decreasing behavior. By using a moving average on the LB over a certain number of iterations $w$, the noise in $\bar{\LB}\br{\lambda} = 1/w \sum_{i=1}^t \hat{\LB}\br{\lambda_{t-i+1}}$ is reduced and $\bar{\LB}$ is stabilized. A typical stopping rule is that terminating the learning after $\bar{\LB}\br{\lambda}$ did not improve for a certain number of iterations, so-called patience parameter ($P$). The final estimate for $\lambda$ is taken as that corresponding to $\max{\bar{\LB}\br{\lambda}}$. Alternatively one might terminate the learning why the change in the parameters is within a certain threshold, this would not even require the computation of the LB, however setting such a threshold can be challenging as it might depend on the scale and length of $\lambda$ \cite{tran2021variational}.

\begin{figure}[htbp]
     \centering
     \includegraphics[width=0.8\textwidth,trim={4cm 0cm 4cm 0cm}]{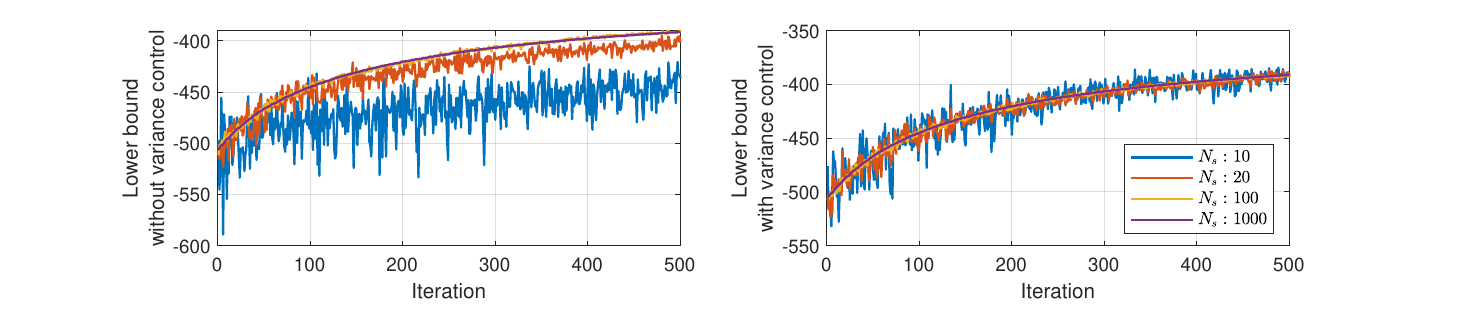}
     \caption{Progression of the $\LB$ without (left) and with (right) control variates, for different numbers of MC samples ($N_s$).}
      \label{fig:cv_LB}
 \end{figure}

\begin{figure}
     \centering
     \includegraphics[width=0.8\textwidth,trim={4cm 0cm 4cm 0cm}]{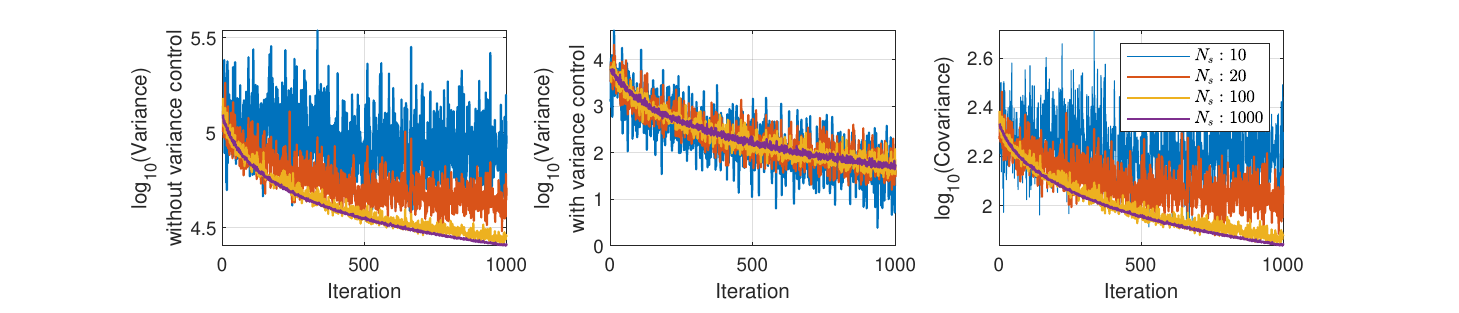}
     \caption{Effect of the use of CVs and number of MC samples ($N_s$) on the variance of the gradients. Top row. $\LB$ without the use of CVs (left) and with CVs (right). Bottom. Variance of the MC gradient for updating $\lam_1$ without CV (left) and with CVs (middle), estimates of the covariance between $\nabla_{m_i}\sbr{\logq} \logpy$ and $\nabla_{m_i}\logq$ (right).}
       \label{fig:cv_grad}
 \end{figure}

\textit{Control variates}. This aspect is discussed in Sec\ref{sec:practical}. Here we discuss the importance of enabling a variance control method by providing empirical illustrations (Credit data). Fig.\ref{fig:cv_LB} shows that the impact of variance control is major under a relatively low number of MC samples. In the left panel, CVs are not used, in the right panel they are: small $N_s$ does lead to improved noise levels in the LB but after CV the progression of the LB and thus of the learning of the variational parameter is (on average) comparable with that of much higher values for $N_s$. On the contrary, without CVs (left panel) the convergence towards the minimum may be slower. The left panel of figure \ref{fig:cv_grad} reports the variance of the MC estimator for the gradient of $\lam_1$ under different $N_s$ schemes without the use of CVs: not surprisingly the higher $N_s$ the lower the variance of the estimator. When enabling for CV, the variance level is on average comparable across different levels of $N_s$ thought deviations can be considerable when $N_s$ is small. The rightmost plot in Fig.\ref{fig:cv_grad} depicts the covariance term in the CV computation \equ\eqref{eq:ci_star} between the score-estimator of the gradient and the variational log-density, upon which depends the effectiveness of the variance reduction.
  Based on these plots we suggest that an appropriate trade-off between gradients' variance, LB smoothness, and computational efforts is about $N_s = 100$, which we adopt throughout our analyses.
  

\textit{Mini-batches}. It is typical in ML applications to work on data subsets or mini-batches. For a batch size $M$ chosen uniformly at random from the total sample of size $N$, the QBVI update can be smoothly implemented on mini-batches by accounting for proper rescaling of $\logpy$ by $N/M$ to obtain an estimate of $\LB$ that is correct in scale. Mini-batch gradients feed the updates of $\mu$ and $\S$ at each iteration resulting in gradient estimates that are more efficient while avoiding computational overhead compared to SGD. On the other hand, mini-batches avoid memory-expensive operations involving the full sample. 
 Table \ref{tab:minibatch} reports estimates and LBs for an experiment involving 50000 records (detail in the caption). The impact of the number of MC draws $N_s$ is aligned with the observations referring to Figure \ref{fig:cv_LB} and \ref{fig:cv_grad}, that is a minor and unbiased effect across batch sizes. Regarding the batch size, we observe some offset with respect to the estimates obtained over the full sample (a single batch). We investigated this effect and found that it turns to be a consequence of a poor approximation of the sample likelihood $\sum_{s=1}^N \log p\br{y\vert \th_s}$ with  $N/M\sum_{s=1}^M \log p\br{y\vert \th_s}$. For small $M$ the variance of $\sum_{i=1}^M \logpy$ is quite considerable across mini-batches which likely overshoots (positively or negatively). Though the minibatch likelihood is on average unbiased, this is rescaled by the additional factors appearing in naive MC estimators (e.g. $\iS_t\br{\theta_s-\mu_t}$ for \equ), and the iterative scheme of the SGD update drives the LB toward different local optima than the ones where it converges under moderate-sized mini-batches. Therefore in applications, we recommend investigating the effect of the minibatch size hyper-parameter with respect to e.g. ML or non-Bayesian optimization point estimation. It has to be noticed that the log-likelihood for independent samples reduces to a summation and that from a practical perspective feed-forwarding the entire data and considering such a sum over it does not represent a computational cornerstone as it generally is in the usual case that involves computing back-propagated gradients, which is certainly more complex to compute and time-consuming.

\begin{table}[htbp]
    \centering
    \caption{Mini-batch experiments. Results for 100 iterations. Synthetic data generated from a logistic regression with true parameters $\beta = (-5,0,-4,-5,2)$. Sample size is 50.000, ML estimates are $\hat{\beta}_{ML}=(-4.56,-0.17,-3.67,-4.97,-1.91)$.\vspace{3pt}}
        \scalebox{0.7}{
\begin{tabular}{crrrrrrrcrrrrr}
      & \multicolumn{6}{c}{Mini-batch size}            &       & \multicolumn{6}{c}{Mini-batch size} \\
      & \multicolumn{1}{c}{64} & \multicolumn{1}{c}{128} & \multicolumn{1}{c}{254} & \multicolumn{1}{c}{514} & \multicolumn{1}{c}{1028} & \multicolumn{1}{c}{2056} &       & 64    & \multicolumn{1}{c}{128} & \multicolumn{1}{c}{254} & \multicolumn{1}{c}{514} & \multicolumn{1}{c}{1028} & \multicolumn{1}{c}{2056} \\
\cmidrule{2-7}\cmidrule{9-14}\multicolumn{7}{c}{$N_s = 25$}                        &       & \multicolumn{6}{c}{$N_s = 50$} \\
$\beta_1$ & -3.03 & -3.34 & -3.73 & -4.35 & -4.57 & -4.37 &       & -3.00 & -3.65 & -3.73 & -4.25 & -4.52 & -4.43 \\
$\beta_2$ & -0.31 & -0.53 & -0.23 & -0.17 & -0.12 & -0.16 &       & -0.39 & -0.53 & -0.18 & -0.17 & -0.12 & -0.17 \\
$\beta_3$ & -2.51 & -2.86 & -3.12 & -3.31 & -3.50 & -3.54 &       & -2.51 & -2.97 & -3.09 & -3.23 & -3.49 & -3.58 \\
$\beta_4$ & -3.12 & -3.45 & -4.18 & -4.40 & -4.76 & -4.78 &       & -3.03 & -3.64 & -4.18 & -4.48 & -4.76 & -4.86 \\
$\beta_5$ & 0.82  & 1.09  & 1.29  & 1.62  & 1.70  & 1.78  &       & 0.80  & 1.09  & 1.41  & 1.56  & 1.71  & 1.82 \\
$\LB$ & -823.16 & -745.56 & -695.26 & -697.95 & -674.96 & -691.86 &       & -819.831 & -743.86 & -694.78 & -697.84 & -674.82 & -691.55 \\
\cmidrule{1-7}\cmidrule{9-14}\multicolumn{7}{c}{$N_s = 100$}                       &       & \multicolumn{6}{c}{$N_s = 200$} \\
$\beta_1$ & -2.91 & -3.61 & -3.78 & -4.31 & -4.51 & -4.34 &       & -2.90 & -3.61 & -3.79 & -4.24 & -4.53 & -4.44 \\
$\beta_2$ & -0.35 & -0.48 & -0.18 & -0.19 & -0.14 & -0.15 &       & -0.33 & -0.47 & -0.18 & -0.18 & -0.12 & -0.16 \\
$\beta_3$ & -2.47 & -2.97 & -3.06 & -3.23 & -3.51 & -3.53 &       & -2.50 & -2.97 & -3.04 & -3.24 & -3.49 & -3.61 \\
$\beta_4$ & -3.03 & -3.64 & -4.13 & -4.42 & -4.77 & -4.79 &       & -3.03 & -3.64 & -4.13 & -4.45 & -4.77 & -4.84 \\
$\beta_5$ & 0.82  & 1.06  & 1.42  & 1.59  & 1.74  & 1.69  &       & 0.84  & 1.05  & 1.43  & 1.58  & 1.71  & 1.75 \\
$\LB$ & -820.58 & -742.72 & -694.84 & -697.26 & -674.24 & -688.18 &       & -819.532 & -741.57 & -694.30 & -697.52 & -674.04 & -686.00 \\
\cmidrule{1-7}\cmidrule{9-14}\end{tabular}%
}
    \label{tab:minibatch}
\end{table}

 \textit{Momentum and adaptive learning rate}. The large variance of the estimate of $g_t = \natgrad LB\br{\lambda_t}$ might be detrimental for the learning process if the learning rate $\beta$ is too large.
 To control the noise of the sample gradient estimates, it is common (e.g. in ADAM or AdaGrad) to smooth the gradient by the use of moving averages
 (momentum method): $\bar{g} = \gamma_g \bar{g} + \br{1-\gamma_g}\hat{g}_t$, with $0< \gamma_g<1$. Also, it is convenient to adaptively decrease the learning rate after a certain number of iterations $t'$. That is, $\lambda_{t+1} = \lambda_t + \beta_t \bar{g}_t$, with e.g. $\beta_t = \min(\epsilon_0,\epsilon_0 \frac{t'}{t})$, for some fixed (small) learning rate $\epsilon_0$.
 
 \textit{Gradient clipping}. Especially at earlier iterations, the norm of $g_t$ can be quite large, and perhaps its variance too if $N_s$ is relatively small. This might lead to updates that are too large and capable of determining a non-positive definite update for $S$, if the above recommendations in \ref{app:Z_Appendix_Positive_definiteness} are ignored. It is a common practice to rescale the $\ell_2$-norm $\vert\vert \cdot \vert\vert$ of the estimated gradient $g_t$ whenever it is larger than a certain threshold $l_{\text{max}}$, while preserving its direction. That is, $g_t$ is replaced by $g_t l_{\text{max}}/\vert\vert g_t \vert\vert$, before momentum is possibly applied \cite{tran2021practical}.

\section{Mean-field QBVI} \label{app:Z_MFQBVI}

Be $\theta = \br{\theta_1, \theta_2, \dots, \theta_k}$. The mean-field variational Bayes framework assumes the following factorization for the variational posterior
 \begin{equation*}
     \q = q_1\br{\theta_1}q_2\br{\theta_2}, \dots, q_k\br{\theta_k} = \prod_{i=1}^k q_k\br{\theta_k}\text{,}
 \end{equation*}
 corresponding to an approximation to the true posterior $p\br{\theta_1,\dots,\theta_k|y}$ by $ \q$, where the dependence between $\theta_1,\dots,\theta_k$ is ignored. 
 
\textit{Example.} Consider a regression problem of a target $y_i$ based on some data $x_i$, formulated as $y_i = f_{\theta}\br{x_i}+\varepsilon_i$, with $\varepsilon_i$ a source of i.i.d. noise. Typically $\varepsilon\sim \mathcal{N}\br{0,\ss}$, with $\ss>0$ corresponding to the residuals' variance. This easily fits the QBVI framework by considering a suitable prior and variational posterior on $\ss$, as done so far for the elements in $\lambda$. Under the mean-field assumption, the variational posterior factorizes as:
 $$
 q\br{\theta,\ss} = q_\lambda\br{\theta} q_\nu\br{\ss} \text{.}
 $$
 The Inverse-Gamma distribution is a common choice for the prior and variational posterior, $p_{\nu_0} =\mathcal{IG}\br{\alpha_0,\beta_0}$, and $q_\nu = \mathcal{IG}\br{\alpha,\beta}$.
 the prior is parametrized over the prior parameter $\nu_0 = (\alpha_0,\beta_0)$ and the variational posterior over $\nu = \br{\alpha,\beta}$, which can be updated with natural gradient updates.
 
 Although the Inverse-Gamma distribution is a member of the exponential family note that $\nu$ is not a natural parameter, but the common scale-location parametrization of the Inverse-Gamma. 
 As opposed to the Gaussian case, in the univariate setting for $\ss$ here discussed, working with $\nu$ is quite practical, second, we aim at providing a worked example of how to adapt mean-field QBVI to such a non-natural parameter, and explicitly use definition \eqref{eq:def_natgrad} for the computation of the FIM in a non-Gaussian case where Proposition \ref{prop:nat_grad_gaussian} is inapplicable. 
 In general, nothing restricts the variational form to differ from the Inverse-gamma specification, and perhaps to update the vector of the natural parameters (within an exponential-family choice of $q_\nu$) via \eqref{eq:update_khan}.
 Following  \eqref{eq:natgrad2mgrad} and \eqref{eq:QBVI_update}, 
 \begin{align}
 \nu_{t+1} = (1-\epsilon)\nu_{t} + \epsilon\br{\nu_0 + \Eq{q} \sbr{\tilde{\nabla}_{\nu} \sbr{\log q_{\nu}\br{\ss}}  \log p\br{y\vert \theta,\ss} }}
 \end{align}
 where $\epsilon$ is the learning rate.  For the Inverse-gamma,  with $\psi'\br{\alpha} = \partial \log \Gamma\br{\alpha} /\partial^2 \alpha$,
 \begin{equation} \label{eq:natgrad_ig}
 \tilde{\nabla}_{\nu}\sbr{\log q_\nu \br{v}} = \iI_{\nu} \nabla_{\nu} \sbr{\log q_\nu\br{v}} = \begin{bmatrix}
           \psi'\br{\alpha} & -\frac{1}{\beta} \\ -\frac{1}{\beta} & \frac{\alpha}{\beta^2}
         \end{bmatrix}
         \begin{bmatrix}
         \log \beta - \frac{\Gamma'\br{\alpha}}{\Gamma\br{\alpha}} - \log v  \\ \frac{\alpha}{\beta}-\frac{1}{v}
         \end{bmatrix}
 \end{equation}
 can be directly plugged into the above two updates for $\alpha_{t+1}$ and $\beta_{t+1}$.
 The lower bound
 \[
 \LB\br{\lambda,\alpha,\beta} = \Eq{\ql}\sbr{\log p\br{\theta_1} + \log p_{\br{\alpha_0,\beta_0}}\br{\theta_2} +\log p_\lambda \br{y\vert \theta_1,\theta_2} - \log q_\lambda \br{\theta_1} - \log q_{\br{\alpha,\beta}}\br{\theta_2}}
 \]
 can be easily evaluated by sampling $\theta_1$ from $\ql$, $\theta_2$ from $q_{\nu}$ based on the values of $\br{\lambda, \alpha,\beta}$ at the current iteration, from which $\log p\br{y\vert \theta_1,\theta_2}$ can be evaluated.  Algorithm \ref{alg:MFQBVI} summarize the above mean-field QBVI approach, applicable for typical univariate regression problems.
 
 In a general setting where the variational posterior is factorized $k$ terms, the above applies to each of the $k$ factors. Whether any of the $q_k$ is a member of the exponential family or nor, 
 one can adopt several alternatives. (i) If $q_k$ is Gaussian, apply the QBVI update discussed in section \ref{subsec:nat_grad_free_gaussian_vi}. (ii-a) If $q_k$ is a member of the exponential family, apply either \eqref{eq:update_khan} or \eqref{eq:LB_gradient_martin} on the natural parameter. (ii-b) If $q_k$ is a member of the exponential family not parametrized over the natural parameter, apply \eqref{eq:update_khan} with the definition \eqref{eq:def_natgrad}. (iii) Abandon natural gradients and adopt the naive Black-box approach \cite{ranganath2014black} on euclidean gradients for the factor $q_k$. (ii-a) corresponds to  is a QBVI update that does not exploit \eqref{eq:natgrad_lambda_grad_m}, from which updates in alternative non-natural parametrizations can be worked out (as actually done for $\zeta=\br{\mu,\iS}$ in \ref{subsec:nat_grad_free_gaussian_vi}). (ii-b) corresponds to the standard VI setting. (iii) is a pure black-box optimizer, that does not exploit the advantages provided by natural gradients.
  
 In case targets and/or one or more of the $q_k$ factors are multivariate, changes in the above discussion are limited to an appropriate choice of the likelihood (and prior-variational posterior pairs).
 
 \textit{Example}. Consider the problem of estimating the parameters of a multivariate normal distribution $p\br{y|\mu,\Sigma} = \mathcal{N}\br{\theta_1 =\mu, \theta_2= \Sigma}$. For the variational posteriors of $\theta_1$ and $\theta_2$ one might respectively take $q_1 \sim \mathcal{N}\br{\mu,S}$ and  $q_2 \sim \text{Wishart}\br{V,1}$, and update the variational parameters $\mu,\iS,V$. The QBVI approach (i) applies for updating $\mu$ and $\iS$, while e.g. (ii-b) for updating $V$.
\begin{minipage}{1\linewidth}
\begin{algorithm}[H]
\caption{Mean-field QBVI for Example B1, diagonal covariance and isotropic prior}\label{alg:MFQBVI}
\begin{algorithmic}[1]
\State \text{Assume: } $q = q_{\mu,S} q_\nu$, $q_{\mu,S} \sim \mathcal{N}\br{\mu,\S}$, $q_\nu \sim \mathcal{IG}\br{\alpha,\beta}$
\State \text{Set hyper-parameters: }$ 0<\epsilon< 1$, $N_s$,
\State \text{Set prior parameters and initial values: } $\tau>0$, $\alpha_0,\beta_0$, $\alpha_1$,$\beta_1$
\State \text{Determine the functions: } $g_\alpha(\ss) = \nabla_\alpha \log q_\nu(\ss),g_\beta(\ss) = \nabla_\beta \log q_\nu(\ss)$
\State \text{Determine the FIM matrix: } $ \iI_{\nu}(\ss) = -\mathbb{E}_{q_\nu}\sbr{\nabla^2_\nu \log q_\nu(\ss)}$
\State \text{Determine the functions: } $(\tilde{g}_\alpha(\ss),\tilde{g}_\beta(\ss))^\top$ = $\iI_\nu(\ss) (g_\alpha(\ss),g_\beta(\ss))^\top$ \Comment{i.e. \eqref{eq:natgrad_ig}}
\State \text{Set:} $t=1$, $\text{Stop} = \texttt{false}$
\While{$\text{Stop} = \texttt{true}$}
\State \text{Generate: } $(\theta,\ss)_{s} \sim q_{t} = q_{\mu_t,S_t} q_{\nu_t}$, $s = 1\dots N_s$
\State \text{Evaluate: } $\hat{g}_S = \tfrac{1}{N_s}\sum_s \br{\iS_t \theta_s - v_t v_t^\top \mu_t} \log p\br{y \vert \theta_s,\ss_s}$
\State \text{Evaluate: } $\hat{g}_\mu = \tfrac{1}{N_s}\sum_s v_t \log p\br{y\vert \theta_s,v_s}$
\State \text{Evaluate: } $\hat{g}_\alpha = \alpha_0 - \alpha_t +\tfrac{1}{N_s}\sum_s \tilde{g}_\alpha (\ss_s) \log p\br{y\vert \theta_s,\ss_s}$ \Comment{ from \eqref{eq:LB_gradient_martin}}
\State \text{Evaluate: } $\hat{g}_\beta =  \beta_0 - \beta_t + \tfrac{1}{N_s}\sum_s \tilde{g}_\beta (\ss_s) \log p\br{y\vert \theta_s,\ss_s}$ \Comment{ from \eqref{eq:LB_gradient_martin}}
\State \text{Update: } $\iS_{t+1} \leftarrow \br{1-\epsilon}\iS_t + \epsilon\sbr{\tau I + \hat{g}_S}$, 
\State \text{Update: } $\mu_{t+1} \leftarrow \mu_t +\epsilon S_{t+1}\sbr{\bm{\tau}_d + \hat{g}_\mu}$
\State \text{Update: } $\alpha_{t+1} \leftarrow \alpha_t + \epsilon \hat{g}_\alpha$ \Comment{\eqref{eq:sgd_nat_update} on $\alpha$}
\State \text{Update: } $\beta_{t+1} \leftarrow \beta_t + \epsilon \hat{g}_\beta$ \Comment{\eqref{eq:sgd_nat_update} on $\beta$}
\State \text{Set: } $t = t+1$, $\text{Stop} = f_{\text{exit}}\br{\cdots}$
\EndWhile
\end{algorithmic}
\end{algorithm}
\end{minipage}

\section{Positive definiteness of the covariance update} \label{app:Z_Appendix_Positive_definiteness}
\subsection{Positive definiteness of the variational covariance matrix}
As of \eqref{eq:update_S_martin}, the update of the covariance matrix $S$ does not grant positive definiteness. For the diagonal Gaussian variational posterior this can be achieved by imposing positivity on the diagonal elements through an appropriate transformation or bounding the learning rate based on the sign and magnitude of the gradient driving the current update. For the full-diagonal case, the $\S$ can be updated on the Gaussian manifold, the methodology advanced in \cite{tran2021variational} and \cite{lin2020handling}. Details on the three approaches are provided in the remainder of this Appendix.

\subsection{Diagonal covariance}
\subsubsection{Explicit constraints}
Assume a posterior diagonal covariance matrix $S$. For a positive definite matrix $S$ its inverse $\iS$ exists and is positive definite as well. As the QBVI update involves $\iS$ rather than $S$ we impose positive definiteness on $\iS$: since we deal with diagonal matrices it suffices to require that all its diagonal entries are positive.
Let $\is_t$ be the column vector of the elements of the diagonal matrix $\iS_t$. To guarantee that $s_t$ remains positive in all its elements \eqref{eq:update_S_diagonal} suggest that at every iteration $t$
\begin{equation*}
 \br{1-\beta}\is_t + \beta\sbr{s^{-1}_0  +\Eq{\ql}\sbr{\br{\is_t-v_t \odot  v_t}\logpy}}>0  
\end{equation*}
needs to hold. Be $h_t$ a short-hand notation for the above term between the square brackets, and for convenience drop the subscript $_t$. The constraint reads
\begin{equation}
\is+\beta\br{h-\is} >0 \text{.} \label{eq:pos_constraint}
\end{equation}
As $0<\beta<1$ the positivity constraint \eqref{eq:pos_constraint} is critical only for the components in $h$ for which $h-\is$ is of negative sign. For these components, a too large value of $\beta$ can cause $\iS$ to be invalid. Eq. \eqref{eq:pos_constraint} rewrites $\beta\br{h-\is} > -\is$ so that
\begin{equation*}
\begin{cases}
\beta > \frac{-\is}{h-\is} &\text{if $h-\is>0$}\text{,}\\
\beta < \frac{-\is}{h-\is} &\text{if $h-\is<0$}\text{.}
\end{cases}
\end{equation*}
The above divisions are intended to be element-wise.
The first condition is irrelevant as it satisfies $\is+\beta\br{h-\is}>0$ by hypothesis, while the second imposes an upper bound on all those components for which $h-\is<0$. Note that $\is+\beta\br{h-\is}>0$ is automatically satisfied for all the $i=1,\dots,d$ components if $\beta$ is smaller than the smallest component of $-\is/\br{h-\is}$. So the constraint:
$$
0<\beta< \argmin_{i:(h_i-\is_i)<0} \frac{-\is_i}{h_i-\is_i} = \beta^\star < 1 \text{.}
$$
In practice, one might set 
$$
\beta = \text{min}\left\lbrace \beta_0, \delta \beta^\star \right\rbrace \text{,}
$$
where $\beta_0$ is a predefined maximum allowed learning rate, and $0<\delta<1$ ensures $\is$ is inside the feasible set \cite{khan2018fastscalable}. With this approach, the diagonal matrix $\iS$ is guaranteed to be positive definite, and $S$ is a valid covariance matrix.

\subsubsection{Transformations}
With $\xi =\xi\br{\lam}$ being an alternative parametrization of the variational density,
$$
\I_\xi = J \I_\lam J^\top, \quad \text{with} \; J=\nabla_\xi \lam
$$
\cite[][Ch.~6.2]{lehmann2006theory}. If $\xi$ is invertible, the Jacobian matrix $J$ is invertible and
$$
\tilde{\nabla}_\xi \LB = \iI\xi \nabla_\xi \LB = \br{J^{{-1}^\top} \iI_\lam J^{-1}} J \br{\nabla_\lam \LB} =\br{\nabla_\lam \xi}^\top \natgrad \LB
$$
where the last equality is in virtue of the Inverse transform theorem [Murray].
Assume $S$ is a diagonal covariance matrix. For some vector $a$, let $a^{-1}$ denote the vector obtained by applying the element-wise division $1/a$, and $\text{diag}\br{a}$ the square diagonal matrix with diagonal $a$. To constrain the $d \times 1$ vector $\is$ corresponding to the diagonal of $\iS$ to be positive, we apply the transform $\xi\,:\,\br{\lam_1^\top,\lam_2^\top}^\top \rightarrow \br{ \lam_1^\top,-\log \br{-\lam_2^\top}}^\top$. Be $\xi^{(2)}=-\log \br{-\lam_2}$, and note that $0<2\exp\br{-\xi^{(2)}} =\is$, guaranteeing that all the entries in $\is$ are positive. 

Under the diagonal covariance assumption $\lam_2$ is the $d\times 1$ vector $-\frac{1}{2}\is$, $\lam_1 = \is \odot \mu$, $\natgrad \LB\br{\lambda}$ is the $2d\times 1$ vector $(\tilde{\nabla}_{\lam_1} \LB\br{\lam}^\top,\tilde{\nabla}_{\lam_2} \LB\br{\lam}^\top)^\top$ and 
\begin{align*}
       \nabla_\lam \xi = \begin{bmatrix}
    I_d & 0\\
    0& -\br{\text{diag}(\lam_2)}^{-1}
    \end{bmatrix} = 
     \begin{bmatrix}
    I_d & 0\\
    0& \text{diag}(2s)
    \end{bmatrix} \text{.}
\end{align*} 
Therefore the natural gradient of the LB w.r.t. $\xi$ becomes
\begin{align*}
    \tilde{\nabla}_\xi \LB\br{\lambda} = \br{\nabla_\lam \xi}^\top \natgrad \LB = \begin{bmatrix}
    \eta_1- \lam_1 + \natgrad \Eq{\ql}\sbr{\logpy} \\
    \text{diag}\br{2s} \br{ \eta_2 -\lam_2+ \natgrad \Eq{\ql}\sbr{\logpy}}
    \end{bmatrix} 
    = \begin{bmatrix}
    \tilde{\nabla}_{\xi^{(1)}} \LB\br{\lam} \\
    \tilde{\nabla}_{\xi^{(2)}} \LB\br{\lam}
    \end{bmatrix}\text{.}
\end{align*}
From \eqref{eq:natgrad2mgrad}, natural gradients can be replaced with euclidean gradients w.r.t. to the expectation parameter $m$. Furthermore, as for \eqref{eq:LB_gradient_martin},  $\nabla_m \Eq{\ql}\sbr{\logpy} =\Eq{\ql}\sbr{ \nabla_m\sbr{\log \ql}\logpy}$, and Proposition \ref{prop:exp_grad_gaussian_solved} applies:
\begin{align*}
 \tilde{\nabla}_{\xi^{(2)}} \LB\br{\lam} 
 &= \text{diag}\br{2s} \br{ \eta_2 -\lam_2+ \natgrad \Eq{\ql}\sbr{\logpy}} \\
 &= 2s \odot  \br{-\frac{1}{2}s^{-1}_0+\frac{1}{2} \is+ \Eq{\ql}\sbr{-\frac{1}{2} \br{ \is_t - v_t\odot v_t}\logpy}}\\
 &= -s\odot \br{s^{-1}_0 - \is +\Eq{\ql}\sbr{ \br{ \is_t - v_t\odot v_t}\logpy}}\text{.}
\end{align*}
This gives the following QBVI update,
\begin{align*}
    \xi^{(2)}_{t+1} &= \xi^{(2)}_{t} -\beta s_t \odot \sbr{s^{-1}_0 - \is_t + \Eq{\ql}\sbr{ \br{ \is_t - v_t\odot v_t}\logpy}}\\
    \is_{t+1} &= \text{diag}\br{2 e^{-\xi^{(2)}_{t+1} }}\\
    \mu_{t+1} &= \mu_t+ \beta s_{t+1}\odot [- \tau \mu_t + \Eq{\ql}\sbr{v_t\logpy}] \text{.}
\end{align*}
\subsection{Full covariance}

\subsubsection{Updating the Cholesky factor}
A $d\times d$ covariance matrix, symmetric and positive definite, requires the specification of $d+\binom{d}{2}$ free parameters, if $S$ is a covariance matrix the Cholesky factor $L$ is the unique lower-triangular matrix such that $S=LL^\top$. While the off-diagonal elements are unconstrained, for $1\leq i \leq d$, the diagonal elements $l_{i,i}$ must be positive. By updating $L$ rather than $S$ \cite{ong2018variational}, a fully unconstrained optimization is achieved by log-transforming the-diagonal and updating the lower-triangular matrix $Z$ where $z_{i,j}= \log l_{i,j}$ for $i=j$, $z_{i,j} = l_{i,j}$ for $i>j$ and$ z_{i,j} = 0$ if $i<j$, and applying the inverse transform to retrieve the Cholesky factor for $S$.  This is an alternative framework that lies outside from the spirit of the QBVI update, as it updates $\zeta=\br{\mu,L}$ instead of $\lam=\br{\lam_1,\lam_2}$, and additionally requires $\iI_\zeta$ and $\nabla_L \log \ql$. We are currently developing and extending the above direction in a separate manuscript.

\subsubsection{Covariance update on the manifold of positive definite symmetric matrices}
\citet{tran2021variational} and \cite{lin2020handling} advanced the idea of performing VI where the variational parameter space is studied as a Riemann manifold. A multivariate Gaussian distribution parametrized by $\zeta=\br{\mu,\S}$ can be viewed as a Riemann manifold equipped with the Riemann metric provided by the Fisher information matrix. In this context, $\iI_\zeta$ approximates as a block-diagonal matrix with blocks $S$, $2S \otimes S$
where $\otimes$ denotes the Kronecker product. This enables to approximate the natural gradients for a Gaussian distribution w.r.t. $\mu$ and $\S$ respectively as $S \nabla_\mu \LB\br{\zeta}$ and $2S \nabla_\S \LB\br{\zeta} S$.
The central insight is that of updating $S$ such that it remains within the (Gaussian) manifold. This is achieved by considering the first-order approximation of the exponential map projecting a point on the tangent space of the manifold at the current point (where the natural gradient lies) back to the manifold, called retraction. The adopted retraction method of \cite{tran2021variational} is justified by \cite{lin2020handling}. \cite{lin2020handling} provides further theoretical and practical developments, especially related to VOGN \cite{osawa_practical_2019}, in the context of the general Bayesian learning rule \cite{khan2021bayesian}. As for the update of the Cholesky factor, this corresponds to an alternative framework that methodologically deviates from QBVI: the positive-definiteness issue constitutes a limitation of QBVI that future work may explore based on the manifold theory.

\section{Experiments, additional results} \label{app:Z_experiments}

\begin{table}[htbp]
    \centering
        \caption{Hyperparameters, apply to all the optimizes listed in Table \ref{tab:par_labour}. The maximum number of iterations is set to 1000. }
    \scalebox{0.9}{
\begin{tabular}{ccccc}
      & Gradient clipping & Learning rate & Patience & Momentum \\
\midrule
Text reference & $l_{\text{max}}$ & $\beta$ & $P$   & $\gamma_g$ \\
Value & 1000  & 0.1   & 500   & 0.4 \\
\midrule
      &       &       &       &  \\
      & Smoothing window & Prior variance & Prior mean & Step adaptive \\
\midrule
Text reference & $w$   & $S_0$ & $\mu_0$ & $t'$ \\
Value & 30    & 5  & 0     & 800 \\
\bottomrule
\end{tabular}%
}

    \label{tab:tab_hyperpar}
\end{table}

The following tables report additional results concerning the variational estimates under different models on the full data.
The VI models include QBVI, QBVI with diagonal covariance matrix (QBVI$^\star$), Black-box Variational Inference (BBVI) \cite{ranganath2014black}, Cholesky Gaussian Variational Bayes (CGVB) \cite{titsias2014doubly}, following the implementation discussed in \cite{tran2021practical}, Manifold Gaussian Variational Bayes (MGVB) \cite{tran2021variational}, Monte Carlo Markov Chain approximation of the actual posterior (MCMC), and Maximum Likelihood (ML). We remark that CGVB is based on the reparametrization trick and constitutes a benchmark that does require models' gradients. MGVB does not, yet it develops over the manifold theory on an approximate-form FIM rather than the exact one based on the duality of the natural- expectation- parameter representations of exponential family distributions. For VI models we include the value of the Lower bound ($\LB$). QBVI regression models under the mean-field framework of Algorithm \ref{alg:MFQBVI} where the regression variance is approximated with an inverse gamma prior and posterior is denoted by (QBVI$\dagger$).
ML variances are extracted from the asymptotic covariance matrix. Models are initialized with the same hyper-parameters.
Throughout the models and datasets, we observe a remarkable alignment between the posteriors' estimates indicating that all the Bayesian models perform comparably, thus the use of our black-box approach seems to come at any cost in terms of bias or increased (posterior) variance of the estimated. Certainly, the smaller number of trainable parameters for the diagonal QBVI version explains the (small) discrepancy observed in the estimated parameters with respect to the alternative optimizers.
It is furthermore interesting to observe that the VI estimates are close to the ML one, indicating that the Gaussian approximation is rather appropriate for the data and models considered. 

Matlab and Python implementations of the QBVI optimizer are available at [TBD] and developed upon the open-source VBLab package, see. \cite{tran2021practical}. 

\begin{table}[htbp]
    \centering
        \caption{Estimates of posteriors' means and variances for the Labour dataset.}
\scalebox{0.85}{
\begin{tabular}{lrrrrrrrrrr}
      & \multicolumn{7}{c}{Posterior means}                   & \multicolumn{3}{c}{Posterior variances} \\
\cmidrule(lr){2-7}  \cmidrule(lr){8-11}      & QBVI$^\star$ & QBVI  & BBVI  & CGVB  & MGVB  & MCMC  & ML    & QBVI  & MCMC  & ML \\
\midrule
$\beta_0$ & 0.672 & 0.675 & 0.679 & 0.678 & 0.678 & 0.676 & 0.679 & 0.041 & 0.040 & 0.040 \\
$\beta_1$ & -1.485 & -1.488 & -1.478 & -1.486 & -1.487 & -1.482 & -1.476 & 0.054 & 0.054 & 0.057 \\
$\beta_2$ & -0.082 & -0.083 & -0.084 & -0.084 & -0.084 & -0.085 & -0.085 & 0.006 & 0.006 & 0.006 \\
$\beta_3$ & -0.572 & -0.574 & -0.572 & -0.574 & -0.574 & -0.566 & -0.571 & 0.015 & 0.015 & 0.014 \\
$\beta_4$ & 0.496 & 0.493 & 0.489 & 0.492 & 0.493 & 0.494 & 0.485 & 0.013 & 0.013 & 0.012 \\
$\beta_5$ & -0.638 & -0.637 & -0.627 & -0.638 & -0.638 & -0.643 & -0.625 & 0.020 & 0.021 & 0.020 \\
$\beta_6$ & 0.608 & 0.609 & 0.602 & 0.609 & 0.609 & 0.612 & 0.599 & 0.020 & 0.020 & 0.020 \\
$\beta_7$ & 0.051 & 0.050 & 0.045 & 0.049 & 0.048 & 0.052 & 0.045 & 0.043 & 0.044 & 0.043 \\
$\LB$    & -358.265 & -356.639 & -359.560 & -356.638 & -356.637 &       &       &       &       &  \\
\bottomrule
\end{tabular}}
    \label{tab:par_labour}
\end{table}

\begin{table}[htbp]
    \centering
    \caption{Estimates of posteriors' means and variances for the German credit data. As an indicative reference, we include the run-time for 10 iterations for the VI models. The current implementation is in Matlab (R2022a) and tested on a low-profile Windows 10 laptop with processor Intel(R) Core(TM) i7-10510U CPU\@ 1.80GHz, 2304 Mhz, 4 cores, 8 logical processors, and 16.0 GB RAM. We report means $\bar{T}$, and standard deviations $\sigma\br{T}$ from the run-time $T$ of 50 independent runs. For MCMC time refers to 1.000 draws (actual analyses use 50.000).}
    \scalebox{0.85}{
\begin{tabular}{lrrrrrrrrrrr}
      & \multicolumn{7}{c}{Posterior means}                   & \multicolumn{3}{c}{Posterior variances} \\
\cmidrule(lr){2-7}    \cmidrule(lr){8-11}   & QBVI$^\star$ & QBVI  & BBVI  & CGVB  & MGVB  & MCMC  & \multicolumn{1}{c}{ML} & QBVI  & MCMC  & ML \\
\midrule
$\beta_0$ & 0.665 & 1.896 & 2.038 & 1.903 & 1.839 & 1.949 & \multicolumn{1}{c}{2.024} & 0.360 & 0.077 & 0.358 \\
$\beta_1$& 0.692 & 0.725 & 0.723 & 0.727 & 0.726 & 0.706 & \multicolumn{1}{c}{0.709} & 0.011 & 0.011 & 0.012 \\
$\beta_2$ & -0.403 & -0.424 & -0.424 & -0.422 & -0.421 & -0.432 & \multicolumn{1}{c}{-0.413} & 0.015 & 0.023 & 0.015 \\
$\beta_3$ & 0.495 & 0.495 & 0.488 & 0.494 & 0.494 & 0.490 & \multicolumn{1}{c}{0.474} & 0.014 & 0.016 & 0.013 \\
$\beta_4$ & -0.152 & -0.154 & -0.152 & -0.154 & -0.155 & -0.151 & \multicolumn{1}{c}{-0.145} & 0.016 & 0.026 & 0.016 \\
$\beta_5$& 0.420 & 0.455 & 0.450 & 0.454 & 0.453 & 0.461 & \multicolumn{1}{c}{0.439} & 0.013 & 0.013 & 0.013 \\
$\beta_6$ & 0.236 & 0.247 & 0.245 & 0.246 & 0.244 & 0.240 & \multicolumn{1}{c}{0.241} & 0.012 & 0.012 & 0.012 \\
$\beta_7$ & 0.211 & 0.199 & 0.196 & 0.199 & 0.198 & 0.200 & \multicolumn{1}{c}{0.190} & 0.010 & 0.010 & 0.010 \\
$\beta_8$ & 0.012 & 0.015 & 0.017 & 0.013 & 0.016 & 0.012 & \multicolumn{1}{c}{0.014} & 0.012 & 0.013 & 0.012 \\
$\beta_9$ & -0.143 & -0.155 & -0.162 & -0.157 & -0.157 & -0.151 & \multicolumn{1}{c}{-0.156} & 0.015 & 0.014 & 0.015 \\
$\beta_{10}$ & 0.218 & 0.179 & 0.171 & 0.180 & 0.181 & 0.162 & \multicolumn{1}{c}{0.163} & 0.014 & 0.014 & 0.014 \\
$\beta_{11}$ & 0.235 & 0.249 & 0.246 & 0.247 & 0.247 & 0.246 & \multicolumn{1}{c}{0.242} & 0.009 & 0.009 & 0.009 \\
$\beta_{12}$ & -0.143 & -0.135 & -0.127 & -0.135 & -0.135 & -0.133 & \multicolumn{1}{c}{-0.128} & 0.012 & 0.014 & 0.012 \\
$\beta_{13}$ & 0.101 & 0.095 & 0.092 & 0.091 & 0.096 & 0.098 & \multicolumn{1}{c}{0.087} & 0.011 & 0.011 & 0.011 \\
$\beta_{14}$ & 0.112 & 0.103 & 0.103 & 0.105 & 0.106 & 0.103 & \multicolumn{1}{c}{0.100} & 0.013 & 0.013 & 0.013 \\
$\beta_{15}$ & 0.457 & 0.444 & 0.411 & 0.447 & 0.444 & 0.421 & \multicolumn{1}{c}{0.396} & 0.029 & 0.040 & 0.029 \\
$\beta_{16}$ & -0.576 & -0.618 & -0.605 & -0.619 & -0.615 & -0.613 & \multicolumn{1}{c}{-0.608} & 0.054 & 0.079 & 0.053 \\
$\beta_{17}$ & 0.905 & 0.818 & 0.771 & 0.822 & 0.813 & 0.798 & \multicolumn{1}{c}{0.768} & 0.141 & 0.104 & 0.140 \\
$\beta_{18}$ & 0.117 & -0.744 & -0.834 & -0.740 & -0.696 & -0.774 & \multicolumn{1}{c}{-0.849} & 0.202 & 0.074 & 0.200 \\
$\beta_{19}$ & -0.221 & -1.028 & -1.185 & -1.035 & -0.980 & -1.072 & \multicolumn{1}{c}{-1.187} & 0.434 & 0.112 & 0.407 \\
$\beta_{20}$ & 0.019 & -0.225 & -0.263 & -0.222 & -0.206 & -0.248 & \multicolumn{1}{c}{-0.265} & 0.176 & 0.086 & 0.174 \\
$\beta_{21}$ & 0.460 & 0.248 & 0.208 & 0.247 & 0.259 & 0.224 & \multicolumn{1}{c}{0.196} & 0.139 & 0.069 & 0.134 \\
$\beta_{22}$ & 0.670 & 0.545 & 0.507 & 0.534 & 0.528 & 0.412 & \multicolumn{1}{c}{0.507} & 0.428 & 0.115 & 0.402 \\
$\beta_{23}$ & 0.285 & 0.102 & 0.069 & 0.095 & 0.095 & 0.076 & \multicolumn{1}{c}{0.064} & 0.142 & 0.089 & 0.136 \\
$\beta_{24}$ & 0.182 & -0.022 & -0.043 & -0.028 & -0.024 & -0.045 & \multicolumn{1}{c}{-0.051} & 0.094 & 0.065 & 0.086 \\
$\LB$ & -421.736 & -414.366 & -415.872 & -414.320 & -414.319 &       &       &       &       &  \\
\midrule
$\bar{T}$      & 0.320 & 0.473 & 0.347 & 0.449 & 0.477 & 0.479  &       &       &       &  \\
 $\sigma\br{T}$     & 0.021 & 0.033 & 0.018 & 0.029 & 0.039 & 0.210  &       &       &       &  \\
\bottomrule
\end{tabular}}
    \label{tab:par_credit}
\end{table}


\begin{table}[htbp]
    \centering
        \caption{Estimates of posteriors' means and variances for the HAR model. For the mean-field models, $\br{\alpha_0,\beta_0} = \br{3,1}$. The reported regression variance $\sigma^2$ for the inverse-gamma models is the one corresponding to the posterior estimates of $\alpha,\beta$.}
\scalebox{0.95}{    
\begin{tabular}{lrrrrrrrrr}
\cmidrule{2-9}      & QBVI $\dagger$ $^\star$ & QBVI $\dagger$ & QBVI $^\star$ & QBVI  & BBVI & CGVB  & MGVB  & ML \\
\midrule
$\beta_0$ & 0.216 & 0.210 & 0.203 & 0.197 & 0.205 & 0.195 & 0.198 & 0.207 \\
$\beta_1$ & 0.490 & 0.497 & 0.482 & 0.513 & 0.489 & 0.518 & 0.516 & 0.494 \\
$\beta_2$ & 0.370 & 0.364 & 0.386 & 0.340 & 0.369 & 0.333 & 0.337 & 0.372 \\
$\beta_3$ & 0.042 & 0.044 & 0.040 & 0.057 & 0.038 & 0.060 & 0.057 & 0.039 \\
$\alpha$ & 30.944 & 13.808 &       &       &       &       &       &  \\
$\beta$ & 4.702 & 4.702 &       &       &       &       &       &  \\
$\sigma^2$ & 0.157 & 0.367 &       &       &       &       &       &  \\
LB    & -455.974 & -466.664 & -3392.743 & -3383.625 & -3378.537 & -3380.095 & -3433.457 &  \\
\bottomrule
\end{tabular}
}
    \label{tab:par_HAR}
\end{table}

\begin{table}[htbp]
    \centering
        \caption{Estimates of posteriors' means and variances for the GARCH(1,1) model. The intercept $\omega$, the autoregressive coefficient of the lag-one squared return $\alpha$ and moving-average coefficient $\beta$ of the lag-one conditional variances need to satisfy the stationarity conditions $\alpha +\beta <1$ and $\omega,\alpha,\beta >0$. Such conditions are unfeasible under a Gaussian approximation, thus we re-parameterize the model. Specifically we estimate the unconstrained parameters $\psi_\omega, \psi_\alpha, \psi_\beta$, where $\omega = f\br{\psi_\omega}, \alpha = f\br{\psi_\alpha}\br{1-f\br{\psi_\beta}}, \beta = f\br{\psi_\alpha}f\br{\psi_\beta}$ with $f\br{x} = \exp\br{x}/\br{1+\exp\br{x}}$ for $x$ real, on which Gaussian's prior-posterior assumptions apply.
    }
\begin{tabular}{lrrrrlrrrr}
      & \multicolumn{4}{c}{Posterior means} &       & \multicolumn{4}{c}{Transformed means} \\
\cmidrule{2-5}\cmidrule{7-10}      & QBVI  & BBVI  & MGVB  & ML    &       & \multicolumn{1}{c}{QBVI} & \multicolumn{1}{c}{BBVI} & \multicolumn{1}{c}{MGVB} & \multicolumn{1}{c}{ML} \\
\midrule
$\psi_\omega$  & -11.930 & -11.989 & -11.874 & -11.874 & \multicolumn{1}{l}{$\omega$} & 0.000 & 0.000 & 0.000 & 0.000 \\
$\psi_\alpha$  & 3.489 & 3.535 & 3.274 & 3.274 & \multicolumn{1}{l}{$\alpha$} & 0.238 & 0.237 & 0.234 & 0.237 \\
$\psi_\beta$ & 1.125 & 1.132 & 1.138 & 1.138 & \multicolumn{1}{l}{$\beta$} & 0.733 & 0.735 & 0.730 & 0.734 \\
$\LB$ & 2564.92 & 2564.29 & 2564.95 &       &       &       &       &       &  \\
\bottomrule
\end{tabular}%
    \label{tab:par_GARCH}
\end{table}

\newpage
\section{Proofs}\label{app:proofs}
\subsection{Natural gradient update}\label{proof:log_trick}

We shall prove that
\begin{equation*}
     \tilde{\nabla}_{\lambda_1} \LB\br{\lambda_1} = \lambda_0-\lambda_1 +  \Eq{q_{\lambda_1}}\br{\tilde{\nabla}_{\lambda_1} \sbr{\logq} \logpy} \text{,}
\end{equation*}
with $\lambda_0$ and $\lambda_1$ being respectively the natural parameters of the prior $p_{\lambda_0}\brt$ and of the variational posterior $q_{\lambda_1}\brt$.
\begin{align}
    \tilde{\nabla}_{\lambda_1} \LB\br{\lambda} &= \tilde{\nabla}_{\lambda_1}  \Eq{q_{\lambda_1}}\sbr{\log \frac{p_{\lambda_0}\brt p\br{y\vert \theta}}{q_{\lambda_1}\brt}} \nonumber\\
   &= \tilde{\nabla}_{\lambda_1}\Eq{q_{\lambda_1}}\sbr{\log \frac{p_{\lambda_0}\brt}{q_{\lambda_1}\brt}} + \tilde{\nabla}_{\lambda_1}\Eq{q_{\lambda_1}}\sbr{\logpy} \label{eq:proofs:d1:1}
\end{align}
Be $p$ and $q$ members of the exponential family. Their densities can be expressed in the form
$$
p_{\lam_0}\brt = h_0\brt \exp\{\phi_0\brt^\top \lam_0 - A_0\br{\lam_0}\}\text{,} \quad \quad q_{ \lam_1}\brt = h_1\brt \exp\{\phi_1\brt^\top \lam_1 - A_1\br{\lam_1}\} \text{,}
$$
with $\phi_0\brt$ and $\phi_1\brt$ the sufficient statistics, $h_0\brt$ and $h_1\brt$ the base functions (or carrying densities) and $A_0$, $A_1$ the log-partition functions of $p$ and $q$ respectively. Let $\langle \cdot \rangle$ denote the dot product. 

We work on the two terms in \eqref{eq:proofs:d1:1} separately. The first term rewrites as 
\begin{align*}
\tilde{\nabla}_{\lambda_1}\Eq{q_{\lambda_1}}\sbr{
   \log \frac{h_0\brt}{h_1\brt} + \phi_0\brt^\top\lambda_0 - \phi_1\brt^\top\lambda_1 - A_0\br{\lambda_0} + A_1\br{\lambda_1}
   }\text{.}
\end{align*}
Thus the derivative with respect to $\lambda_1$ is
\begin{align*}
  \frac{\partial}{\partial \lambda_1}  \Eq{q_{\lambda_1}}\sbr{\log \frac{h_0\brt}{h_1\brt}} + \frac{\partial}{\partial \lambda_1} \Eq{q_{\lambda_1}}\left[ \phi_0\brt^\top \lambda_0 \right] 
   - \frac{\partial}{\partial \lambda_1} \langle m_1,\lambda_1 \rangle + m_1 \text{,}
\end{align*}
as $m_1 = \Eq{q_{\lambda_1}} \sbr{ \phi_1\brt}$ and $m_1 = \partial A\br{\lambda_1} /\partial \lambda_1$ too.

By expanding the derivative of the dot product, 
\begin{align*}
    \frac{\partial}{\partial \lambda_1}  \Eq{q_{\lambda_1}}\sbr{\log \frac{h_0\brt}{h_1\brt}} &+
    \frac{\partial}{\partial \lambda_1} \Eq{q_{\lambda_1}}\left[ \phi_0\brt  \right]^\top \lambda_0 
    -\langle \frac{\partial}{\partial \lambda_1}m_1,\lambda_1 \rangle \text{.}
\end{align*}
and recalling the exponential family property
$$\I_{\lambda_1} = \frac{\partial^2}{\partial \lambda_1  \partial \lambda_1} A\br{\lambda_1} = \frac{\partial}{\partial \lambda_1} m_1 \text{,}$$
we lastly have
\begin{align*}
\tilde{\nabla}_{\lambda_1}\Eq{q_{\lambda_1}}\sbr{\log \frac{p_{\lam_0}\brt}{q_{\lambda_1}\br{\theta}}} 
&=\iI_{\lambda_1} \sbr{ \frac{\partial}{\partial \lambda_1}  \Eq{q_{\lambda_1}}\sbr{\log \frac{h_0\brt}{h_1\brt}}+ \frac{\partial}{\partial \lambda_1} \Eq{q_{\lambda_1}}\left[ \phi_0\brt  \right]^\top \lambda_0 - \I_{\lam_1} \lambda_1}\\
&= \iI_{\lambda_1} \frac{\partial}{\partial \lambda_1} \Eq{q_{\lambda_1}}\sbr{\log \frac{h_0\brt}{h_1\brt}}  + \iI_{\lambda_1} \frac{\partial}{\partial \lambda_1} \Eq{q_{\lambda_1}}\left[ \phi_0\brt  \right]^\top \lambda_0 -  \lambda_1 \text{.}
\end{align*}
If $p_{\lambda_0}$ and $q_{\lambda_1}$ are exponential family members of the \textit{same} parametric form, they share the sufficient statistic and base functions, i.e. $\phi_0 = \phi_1$ and $h_0 = h_1$. Then,
$$
 \frac{\partial}{\partial \lambda_1} \Eq{q_{\lambda_1}}\left[ \phi_0\brt  \right]  = \frac{\partial}{\partial \lambda_1} \Eq{q_{\lambda_1}}\left[ \phi_1\brt  \right] = \frac{\partial}{\partial \lambda_1} m_1  = \I_{\lambda_1}\text{,} 
$$
thus the very simple form:
$$
\tilde{\nabla}_{\lambda_1}\Eq{q_{\lambda_1}}\sbr{\log \frac{p_{\lambda_0}\brt}{q_{\lambda_1}\brt}} = \lambda_0 - \lambda_1 \text{.}
$$
For the second term, 
\begin{align*}
    \tilde{\nabla}_{\lambda_1}  \Eq{q_{\lambda_1}}\sbr{\logpy} &= \int \tilde{\nabla}_{\lambda_1} \sbr{ q_{\lambda_1}\brt \logpy} d\theta\\
    &= \int \tilde{\nabla}_{\lambda_1} \sbr{q_{\lambda_1}\brt} \logpy + q_{\lambda_1}\brt \tilde{\nabla}_{\lambda_1} \sbr{\logpy} d\theta\\
    & = \int q_{\lambda_1}\brt \tilde{\nabla}_{\lambda_1}  \sbr{\log q_{\lambda_1}\brt} \logpy\, d\theta + 0 \\
    &= \Eq{q_{\lambda_1}}\sbr{\tilde{\nabla}_{\lambda_1}  \sbr{\log q_{\lambda_1}\brt} \logpy}
\end{align*}
The score function exists, scores and likelihoods are bounded, and via Theorem of dominated convergence derivatives and integrals can be exchanged \cite{ccinlar2011probability}.

\subsection{Proposition \ref{prop:exp_grad_gaussian_solved}: gradient with respect to the expectation parameter}\label{proof:grad_wrt_m}
We shall prove that under $\qlt\sim \mathcal{N}\br{\mu,S}$, with $m_1 = \mu$, $m_2 = \mu \mu^\top +S$ and $V = \br{\theta -\mu}\br{\theta -\mu}^\top$ it holds that
\begin{align*}
    &\nabla_{m_1} \log \qlt  = \iS\br{\theta -V \iS\mu} \text{,}\\
    &\nabla_{m_2} \log \qlt  = -\frac{1}{2}\br{\iS - \iS V \iS} \text{.}
\end{align*}
From Proposition \ref{prop:nat_grad_gaussian}, gradients of $\log \qlt$ with respect to $S$ and $\mu$ are required.
An solution to the rather complex problem $\nabla_S \log\qlt$ is provided in \cite{anderson1985maximum} (indeed standard maximum likelihood estimation simplifies the problem by considering the much simpler derivative with respect to $\iS$).

For the quadratic form in $\log \qlt$, one has $\nabla_\mu \sbr{ \br{\theta -\mu}\iS\br{\theta -\mu}^\top} = 2\iS \br{\theta -\mu}$, therefore
\begin{align}
    \nabla_{\mu} \log \qlt = \iS \br{\theta-\mu} \text{.}
\end{align}
Form Proposition \ref{prop:nat_grad_gaussian},
\begin{align*}
   \nabla_{m_1} \log \qlt &=  \nabla_\mu \log \qlt -2\sbr{\nabla_S \log \qlt} \mu\\
   &=\iS\br{\theta -\mu} + \br{\iS - \iS V \iS}\mu
   =\iS \br{\theta-V\iS\mu}\text{,}\\
   \nabla_{m_2} \log \qlt &= \nabla_S \log\qlt \text{.}
\end{align*}

\subsection{QBVI under a full-covariance Gaussian variational posterior}\label{proof:QBVI}
Update for Gaussian variational posterior in the natural parameter space.

Apply the SGD update $\eqref{eq:sgd_nat_update}$ for the loss $\LB$ 
$$
\natgrad \LB\br{\lambda} = \eta - \lambda +\Eq{\ql}\br{\nabla_m\sbr{\log \qlt} \logpy}
$$
with the natural gradient in \eqref{eq:LB_gradient_martin} to obtain the generic update in the natural parameters space:
\begin{align*}
     \lambda_{t+1} &= \lambda_t + \beta\sbr{\eta - \lambda_t + \Eq{\ql}\sbr{\natgrad\sbr{\log \qlt} \logpy}}\\
    & = \br{1-\beta}\lambda_{t} + \beta\sbr{\eta +\Eq{\ql}\sbr{\natgrad\sbr{\log \qlt} \logpy}}
\end{align*}
As $\tilde{\nabla}_{\lambda} \sbr{\log \qlt}= \nabla_m\sbr{\log \qlt}$,  we replace the natural gradients with the gradients with respect to $m$ given in Proposition \ref{prop:exp_grad_gaussian_solved},
\begin{align*}
 \Eq{\ql}\sbr{\tilde{\nabla}_{\lambda_1}\sbr{\logq} \logpy}
  &= \Eq{\ql}\sbr{\iS\br{\theta- V \iS \mu} \logpy}\\
 & = \iS\Eq{\ql}\sbr{\br{\theta- V \iS \mu} \logpy}\text{,}
 \end{align*}
 \begin{align*}
 \Eq{\ql}\sbr{\tilde{\nabla}_{\lambda_2}\sbr{\logq} \logpy}
 &= -\frac{1}{2} \Eq{\ql}\sbr{\br{\iS-\iS V \iS } \logpy}\\
 & = -\frac{1}{2} \iS\Eq{\ql}\sbr{\br{I-V \iS} \logpy} \text{.}
\end{align*}
The updates in the natural parameter space results in
\begin{align} 
    \lambda^{(1)}_{t+1} &= \br{1-\beta} \lambda^{(1)}_{t} + \beta \eta_t^{(1)}  
     \beta\Eq{\ql}\sbr{\iS_t\br{\theta-V_t\iS_t\mu_t}\logpy} \text{,}\label{eq:bayes_update_lambda_1} \\
    \lambda^{(2)}_{t+1} &= \br{1-\beta} \lambda^{(2)}_{t} + \beta \eta_t^{(2)} 
    -\frac{1}{2}\beta\Eq{\ql}\sbr{\br{\iS_t-\iS_tV_t\iS_t} \logpy} \label{eq:bayes_update_lambda_2} \text{,}
\end{align}
where $\lambda^{(1)}$, $\lambda^{(2)}$ ($\eta^{(1)}$, $\eta^{(2)}$) are respectively the first and second natural parameters of the variational Gaussian posterior (Gaussian prior) and $I$ is the  identity matrix of appropriate size. The learning rate $\beta$ can defined component-wise $\beta = \br{\beta^{(1)},\beta^{(2)}}^\top$ and be adaptive $\beta = \beta_t$.

Update for $\iS$. In \eqref{eq:bayes_update_lambda_2} substitute $\lambda_2  = -\frac{1}{2}\iS$:
\begin{align}
    -\frac{1}{2}\iS_{t+1} &= -\frac{1}{2}\br{1-\beta}\iS_t 
    + \beta \sbr{-\frac{1}{2}\iS_0 + \Eq{\ql}\sbr{\nabla_{m_2} \logq} \logpy}\label{eq:update_S_first_line}\\
    \iS_{t+1} &= \br{1-\beta}\iS_t +\beta \iS_0 -2\beta\sbr{\Eq{\ql}\sbr{\nabla_{m_2} \logq} \logpy}\nonumber\\
    \iS_{t+1} &= \br{1-\beta}\iS_t\nonumber
    +\beta \sbr{\iS_0  + \Eq{\ql}\sbr{\br{\iS_t-\iS_tV_t\iS_t}\logpy}}\nonumber
\end{align}

Update for $\mu$. In \eqref{eq:bayes_update_lambda_1} substitute $\lambda_1 = \iS \mu$:
\begin{align*}
    \mu_{t+1} &= \br{1-\beta}\S_{t+1}\iS_t \mu_t + \beta \S_{t+1}[\iS_0\mu_0 + \Eq{\ql}\sbr{ \nabla_{m_1}\sbr{\log \qlt}\logpy}] \text{.}
\end{align*}

Multiply the update for $S_{t+1}$ in \eqref{eq:update_S_first_line} by $\S_{t+1}\mu_t$ to obtain
\begin{align*}
    \br{1-\beta}\S_{t+1}\iS_t \mu_t = \mu_t - \beta\S_{t+1}\br{\iS_0 \mu_t - 2\sbr{\Eq{\ql}\sbr{\nabla_{m_2} \logq} \logpy} \mu}
\end{align*}

So for update of $\mu_{t+1}$, using the first relation from Proposition \ref{prop:nat_grad_gaussian},
\begin{align*}
    \mu_{t+1} &= \mu_t + \beta\S_{t+1}[\iS_0\mu_0 - \iS_0 \mu_t\Eq{\ql}\sbr{ \nabla_{m_1}\sbr{\log \qlt}\logpy}\\&+2\Eq{\ql}\sbr{\nabla_{m_2} \sbr{\logq} \logpy} \mu]\\
    &= \mu_t + \beta\S_{t+1}[\iS_0\mu_0 - \iS_0 \mu_t + \Eq{\ql}\sbr{ \nabla_{\mu}\sbr{\log \qlt}\logpy}]\\
    &= \mu_t + \beta\S_{t+1}[\iS_0\mu_0 - \iS_0 \mu_t + \Eq{\ql}\sbr{\iS_t\br{\theta-\mu_t}\logpy}]
\end{align*}

\subsection{Control variates (i). Validity of the control variate}\label{proof:cv_valid}
For $\nabla_{m_i}\sbr{\log q\br{\theta}}$ to be valid control variate it must hold $\Eq{q}\sbr{\nabla_{m_i}\log q\br{\theta}}=0$. Let the $j$-th component of $m_i$ be denoted by $m^{(j)}_i$. It suffices to prove that  $\Eq{q}\sbr{\nabla_{m^{(j)}_i}\log q\br{\theta}}=0$, $\forall{j}$.
\begin{align*}
    \Eq{q}\sbr{\nabla_{m^{(j)}_i}\logq} &=  \Eq{q}\sbr{\frac{\frac{\partial }{\partial {m^{(j)}_i}}\q}{q\br{\theta} }}\\
    &= \int q\br{\theta} \frac{\frac{\partial}{\partial m^{(j)}_i} \q }\q d\theta 
    = \frac{\partial}{\partial m^{(j)}_i} \int \q d\theta
    = \frac{\partial}{\partial m^{(j)}_i} 1 = 0 \text{,} \quad \forall{j} \text{.}
\end{align*}
If $m_i$ is a matrix, the above analogously applies to $\vec\br{m_i}$, and the corresponding matrix of derivatives is the zero matrix.

\subsection{Control variates (ii). Variance of the gradients} \label{proof:cv_var}
Recall the gradients with respect to the expectation parameters can be written in terms of gradients with respect to the mean and variance of the Gaussian variational posterior $\theta \sim q\br{\mu,\S}$, with $\mu \in \R^{K\times 1}$ begin the column vector of means, $\S \in \R^{K\times K}$ the variance-covariance matrix. 
From Proposition \ref{prop:exp_grad_gaussian_solved} recall that
\begin{align*}
\nabla_{m_1} \logq &= \nabla_{\mu} \logq -2 \nabla_{\S} \sbr{\logq}\mu  \text{,}\\
\nabla_{m_2} \logq &= \nabla_{\S} \logq \text{,}
\end{align*}
with $m_1 = \mu$ and $m_2 = \mu \mu^\top +\S$.
This implies that three terms are required for computing the variance of each gradient, these are: (i) $\Var\br{\nabla_{\mu} \logq }$, (ii) $\Var\br{\nabla_{\S} \logq }$, and (iii) $\Cov\br{\nabla_{\mu} \logq, \nabla_{\S} \logq}$.

\textit{Notation.}
For a $K \times 1$ vector $a$, $\vcov\br{a}$ is the usual variance-covariance matrix  $\mathbb{E}[\br{a-\E{a}}\br{a-\E{a}}^\top]$.
For a $K\times L$ matrix $A$, by $\Var\br{A}$ we mean the $K\times L$ matrix of the variances of the individual elements in $A$, that is $\br{\Var\br{A}}_{ij} = \Var\br{A_{ij}}$.
$\vcov\br{A}$ is the matrix size $K^2 \times K^2$ defined as the variance-covariance matrix of the vector $\vec[A]$, i.e. $\vcov(A) = \vcov\br{\vec[A]}$. In the univariate case, $\Cov$ is used in place of $\vcov$ to simplify the notation.



\textbf{(i) Variance of the gradient of the variational likelihood with respect to $m_2$.}

It is useful to recall the relationship between the Wishart and the multivariate Gaussian distribution, and a further result on the linear combinations of Wishart distributions through non-random matrices.

\begin{prop_nat_grad_gaussian}\label{prop:wishard}
Let $X_1,\dots, X_n$ be $n$ independent $K \times 1$ random vectors all having a multivariate normal distribution with mean zero and covariance matrix $\frac{1}{n}S$. Let $K \leq n$. Define $W = \sum X_i X_i^\top$, then $W$ has a Wishart distribution with parameters $n$ and $S$, denoted by $W\br{n,S}$. For a proof see \cite{ghosh2002simple}.
\end{prop_nat_grad_gaussian}

\newtheorem{theorem}{Theorem}
\begin{theorem}\label{th:MSM}
If $A$ is a $W\br{n,\Sigma}$ and $M$ is $k\times m$ of rank $k$ then $MAM'$ is $W\br{n,M\Sigma M}$. For a proof see \cite{muirhead2009aspects}.
\end{theorem} 

\newtheorem{cor}{Corollary}
\begin{cor}\label{cor:wishart_normal}
Be $V=\br{X-\mu}\br{X-\mu}^\top$ with $X\sim N\br{\mu,\S}$, then $\iS V \iS \sim W\br{1,\iS}$.
\end{cor}
\textit{Proof.} $\br{X-\mu}$ distributes as a $K$-variate normal distribution with mean zero and covariance matrix $\S$. Following Proposition \ref{prop:wishard}, $V = \br{X-\mu}\br{X-\mu}^\top \sim W\br{1,\S}$. $\iS$ is symmetric and of rank $K$, thus by applying Theorem \ref{th:MSM} one has $\iS V \iS \sim W\br{1,\iS}$.

For the covariance matrix of $\nabla_{m_2} \log \qlt$ we have
\begin{align*}
\vcov\br{\nabla_{m_2} \logq} &= \vcov(-\frac{1}{2} \iS +\frac{1}{2} \iS V \iS)= \frac{1}{4}\vcov\br{\iS V \iS}\text{.}
\end{align*}
By Corollary \ref{cor:wishart_normal} the covariance matrix of $\iS V \iS$ is that of the $W\br{1,\iS}$ distribution, indicated by $Q$, then
$$
\vcov\br{\nabla_{m_2} \logq} = \frac{1}{4}Q
$$
By definition, $Q=\vcov\br{\text{vec} \br{\iS V \iS}}$, therefore the variances of the individual entries in $\iS V \iS$ are found on the diagonal of $Q$. That is $\text{diag}\br{Q} = \text{vec} \br{\Var\br{\iS V \iS}}$, so 
\begin{align*}
    \Var\br{\nabla_{m_2} \logq} = \frac{1}{4}\text{vec}^{-1} [\text{diag}\br{Q}] \text{.}
\end{align*}
For the Wishart distribution, the term $\text{diag}\br{Q}$ can be easily computed as 
$$\iS \odot \iS + \text{diag}\br{\iS}\text{diag}\br{\iS}^\top\text{.}$$

\textbf{(ii) Variance of the gradient of the variational likelihood with respect to $m_1$}.

We prove the following:
$$\vcov\br{\nabla_{m_1}\logq} = \iS \br{S + D} {\iS} \, \text{,}$$ 
where $D$ is defined in \eqref{eq:kron_Vz}.

From Proposition \ref{prop:exp_grad_gaussian_solved}, 
\begin{align*}
 \nabla_{m_1} \logq &= \iS\br{\theta-\mu} +  \br{\iS - \iS V \iS}\mu\\
 & = \iS\br{\theta - V \iS \mu}\\
 & =\iS\br{\theta -V z} \text{,}
\end{align*}
with $z = \iS \mu$ being a constant column vector and $V = \br{\theta-\mu}\br{\theta-\mu}^\top$.
The covariance matrix corresponds to,
\begin{align}\label{eq:cov_grad_m1}
\vcov\br{ \nabla_{m_1} \logq} &= \iS \vcov\br{\theta - Vz}{\iS}^\top \\
&= \iS \br{\vcov\br{\theta} + \vcov\br{Vz} + \vcov\br{\theta,Vz}} {\iS}  \text{.}\nonumber
\end{align}
Since $\vcov\br{\theta}$ is trivially equal to $\S$, it turns out that there are two terms that need to be addressed. \\

\textit{(ii-a) Term $\vcov\br{Vz}$.}

We first develop on the $K\times K$ matrix $\vcov\br{Vz}$. The diagonal elements correspond to variances, that is for $j=1,\dots,K$,
\begin{align*}
    \Var[\br{Vz}_{jj}] = \sum_{i=1}^V z_i^2\Var\br{V_{ji}} +2\sum_{i \neq h}z_i z_j\Cov\br{V_{ji},V_{jh}} \text{.}
\end{align*}
For any $j$ all the relevant variance and covariance terms are found in the variance-covariance matrix $Q$ of the $W\br{1,S}$ distribution of $V$. In particular, for $j = 1$ the relevant part of $Q$ is the sub-matrix  $Q^{(1,1)}$ extracted from $Q$ by taking rows $1,\dots, K$ and columns $1, \dots, K$
\begin{align*}
    \Var[\br{Vz}_{11}] = \sum_{i=1}^V z_i^2Q^{(1,1)}_{ii} +2\sum_{i \neq j}z_i z_j Q^{(1,1)}_{ij} \text{,}
\end{align*}
for $i,j = 1,\dots, K$. For $j=2$, $Q^{(2)}$ is extracted from $Q$ by taking rows and columns from $K+1,\cdots,2K$, and similarly
\begin{align*}
    \Var[\br{Vz}_{22}] = \sum_{i=1}^V z_i^2Q^{(2,2)}_{ii} +2\sum_{i \neq j}z_i z_j Q^{(2,2)}_{ij}  \text{,}
\end{align*}
again, for $i,j = 1,\dots, K$. Analogously for the $j$th row, $Q^{(j,j)}$ is extracted from $Q$ by taking rows and columns from $(j-1)K+1,\cdots,jK$, and similarly
\begin{align*}
    \Var[\br{Vz}_{jj}] = \sum_{i=1}^V z_i^2Q^{(j,j)}_{ii} +2\sum_{i \neq j}z_i z_j Q^{(j,j)}_{ij}  \text{,}
\end{align*}
for $i,j = 1,\dots, K$. The $j$-th variance can be analogously expressed in terms of matrix multiplication as
\begin{align*}
    \Var[\br{Vz}_{jj}] = z^\top Q^{(j,j)} z \text{,}
\end{align*}
which can be proved by expanding the matrix product and observing that $Q^{(j,j)}$ is symmetric. 
For the generic covariance term $\Cov(\br{Vz}_{i},\br{Vz}_j)$, one has
\begin{align*}
    \Cov\br{\br{Vz}_{i},\br{Vz}_j} &= \Cov\br{\sum_h V_{ih}z_h,\sum_k V_{ik}z_k}
    = \sum_h \sum_k z_h z_k \Cov\br{V_{ih},V_{ik}} \text{.}
\end{align*}
Again, the relevant covariance terms are found in $Q$. Be $Q^{(i,j)}$ the sub-matrix of $Q$ obtained by extracting rows 
$(i-1)K+1,\cdots,iK$ and columns $(j-1)K+1,\cdots,jK$. Similarly to the variance case
\begin{align} \label{eq:vcov_general_term}
    \Cov\br{\br{Vz}_{i},\br{Vz}_j} = z^\top Q^{(i,j)}z \text{,}
\end{align}
That is, the generic $i$-th row of $\vcov\br{Vz}$ corresponds to the vector
\begin{align*}
    \br{z^\top Q^{(i,1)}z,z^\top Q^{(i,2)}z,\dots, z^\top Q^{(i,K)}z} \text{.}
\end{align*}
The partitioned matrix $Q$ of size $K^2 \times K^2$ into the $Q^{(i,j)}$ sub-matrices each of size $K \times K$ is vectorized into a $K^3 \times K$ matrix of $K^2$ vertically-stacked blocks of $K\times K$ matrices and further block-diagonalized to obtain the $K^3 \times K^3$ matrix  
\begin{align*}
BQ &= \text{Bdiag}(Q^{(1,1)},Q^{(2,1)}, \dots, Q^{(K,1)}, Q^{(1,2)},\dots , \\
 & Q^{(K,2)}, Q^{(1,3)}, \; \dots \;, Q^{(K,K-1)}, Q^{(1,K)},\dots,  Q^{(K,K)} ) \text{.}
\end{align*}
In this way, a compact form for the whole variance-covariance matrix
can be retrieved in terms of Kronecker ($\otimes$)  products as
\begin{align}\label{eq:kron_Vz}
D=\text{vec}^{-1}\sbr{\text{diag}\br{\br{I_{K \times K} \otimes z}^\top BQ \br{I_{K \times K} \otimes z}}} \text{,}
\end{align}
where $\text{diag}$ is the operator that extracts the diagonal elements of a matrix into a column vector, $I_{K \times K}$ denotes the identity matrix of size $K \times K$ and ${\text{vec}}^{-1}$ the inverse of the vectorization operator.
Eq. \eqref{eq:kron_Vz}, can be proved by expanding the products and recognizing that the product of the three matrices corresponds to a diagonal matrix whose diagonal is equal to $\text{vec}[\vcov(Vz)]$, thus the composed function $\text{vec}^{-1}[\text{diag}(\cdot)]$.
Eq. \eqref{eq:kron_Vz} provides a compact notation and formal method for computing $\vcov(Vz)$. Thought the $BQ$ matrix is sparse and the matrix products in \eqref{eq:kron_Vz} are computationally efficient, the initialization of $BQ$ requires a $N^3 \times K^3$ array which is likely to exceed the maximum array size, thus in practical applications one might construct $\vcov(Vz)$ from \eqref{eq:vcov_general_term} by exploiting the symmetry of $\vcov(Vz)$, which however leads to $\mathcal{O}\br{K^2}$ complexity.\\

\textit{(ii-b) Term $\vcov\br{\theta,Vz}$}

The second element of interest are the covariances $\Cov(\theta_j,\br{Vz}_j)$, that is, the pair-wise covariances between the rows of the column-vectors $\theta$ and $Vz$,
\begin{align}\label{eq:cov_theta_Vz}
    \Cov(\theta_j,\br{Vz}_j) &= \Cov\br{\theta_j, V_{j1}z_1 + \dots + V_{jK}z_k} \nonumber \\
    &= \Cov\br{\theta_j,V_{jj}z_j} + \sum_{i \neq j} \Cov\br{\theta_j,V_{ji}z_i} \text{.}
\end{align}
Regarding the first term in the above sum, it is useful recalling that for a standard normal $Y$, $\Cov\br{Y,Y^2} = 0$, from which
\begin{align*}
 \Cov\br{\theta_j,\theta_j^2} &= \Cov\br{\mu_j +S_{jj} Y,\br{\mu_j +S_{jj} Y}^2}
 = S_{jj}^3 \Cov\br{Y,Y^2} + 2\mu_j \S_{jj}^2 = 2\mu_j\S_{jj}\text{.}
\end{align*}
Therefore,
\begin{align}\label{eq:cov_theta_Vz_jj}
\Cov&\br{\theta_j,V_{jj}z_j}  = z_j\Cov\br{\theta_j,\br{\theta_j-\mu_j}^2}   
= z_j\br{\Cov\br{\theta_j,\theta_j^2} -2\mu_j\S_{jj}} = 0\text{.} 
\end{align}
For the generic term $\Cov\br{\theta_j,V_{ji}z_i}$,
\begin{align}
     z_i\Cov\br{\theta_{j},\br{\theta_j-\mu_j}\br{\theta_i-\mu_i}} &= z_i\br{\cov{\theta_j,\theta_j\theta_i}-\mu_i \S_{jj} + \mu_j S_{ji}} \nonumber \\
    & = z_i\br{\E{\theta_j^2\theta_i}-\E{\theta_j\theta_i}\E{\theta_i}-\mu_i \S_{jj} + \mu_j S_{ji}}\text{.}  \label{eq:cv_two_expectations} 
\end{align}
As the variational distribution for $\theta$ is a $K$-variate Gaussian, standard marginalization and conditioning results imply that $\theta_i \theta_j$ are jointly Gaussian and $\theta_j\vert\theta_i$ is a conditional univariate Gaussian distribution with mean $\mu_j +S_{ji}\iS_{ii}\br{\theta_i -\mu_i}$ and variance $\iS_{jj}-S_{ji}\iS_{ii}\S_{ij}$.
By the law of the total expectation, for the second term in \eqref{eq:cv_two_expectations} we have
\begin{align*}
    \mathbb{E}\left[\theta_i \E{\theta_j\vert\theta_i}\right]\E{\theta_j} &= \E{\theta_i\br{\mu_j +S_{ji}\iS_{ii}\br{\theta_i -\mu_i}}}\mu_i\\
    &=\E{\theta_i \mu_j + \theta_i \S_{ji}\iS_{ii}\br{\theta_i - \mu_i}}\mu_j\\
    &=\br{\mu_i \mu_j + \S_{ji}\iS_{ii}\E{\theta_i^2} -\S_{ji}\iS_{ii} \mu_i^2}\mu_j\text{.}
\end{align*}
For $Y$ being a univariate Gaussian of mean $\mu$ and unit variance, $Y^2$ distributes as a non-central chi-squared with one degree of freedom and centrality parameter $\lambda = \mu^2$, for which the mean is $1+\lambda$ and the variance $2+4\lambda$.
Similarly, $\theta_i/S_{ii}$ follows a non-central chi-squared with one degree of freedom and centrality parameter $\lambda = \mu_i^2/ \S_{ii}$. So $\mathbb{E}[ \theta_j^2] = (1+\frac{\mu_i^2}{S_{ii}})S_{ii} = S_{ii}+\mu_i^2 $, and lastly
\begin{align}\label{eq:expectation_term_2}
    \E{\theta_j\theta_i}\E{\theta_i} = \mu_i \mu_j^2 + \S_{ji}\mu_j \text{.}
\end{align}
Also for the first expectation in \eqref{eq:cv_two_expectations}, we write $\mathbb{E}[\theta_j^2\theta_i] = \mathbb{E}[\theta_i \mathbb{E}[\theta_j^2\vert \theta_i]]$ and recognize that $\theta_j^2 \vert \theta_i$ is also a non-central chi-squared, so that
\begin{align*}
    \mathbb{E}[\theta_j^2\vert \theta_i] &= S_{j\vert i}+\mu_{j\vert i}^2\\
    &=\S_{j\vert i} + \br{\mu_j + \S_{ji}\iS_{ii}\br{\theta_i - \mu_i}}^2\\
    &=\S_{j\vert i}
    +[\mu_j^2 + 2\mu_j \S_{ji}\iS_{ii}\br{\theta_i - \mu_i}
     +\S^2_{ji}\S^{-2}_{ii}\br{\theta_i^2 -2\theta_i \mu_i + \mu_i^2}]\text{.}
\end{align*}
Now $\mathbb{E}[\theta_j^2\theta_i]$ can be expanded as
\begin{align}\label{eq:expectation_squared}
   \mathbb{E}[\theta_i \mathbb{E}[\theta_j^2\vert \theta_i]] &=  \mathbb{E}[ \theta_i \S_{j\vert i} + \theta_i \mu^2_j + 2\mu_j \S_{ji}\iS_{ii} \theta_i^2 \nonumber \\ 
   &- 2\mu_j \mu_i \S_{ji}\iS_{ii} \theta_i  + \S^2_{ji}\S^{-2}_{ii} \br{\theta_i^3 -2\mu_i\theta_i^2 +\mu_i^3}]\nonumber \\ 
   &= \mu_i S_{j\vert i} + \mu_i \mu_j^2 + 2\mu_j \S_{ji}\iS_{ii}\E{\theta_i^2}\nonumber \\ 
   &-2\mu_j \mu_i^2 \S_{ji}\iS_{ii} +\S^2_{ji}\S^{-2}_{ii}\sbr{\E{\theta_i^3} - 2\mu_i\E{\theta_i^2} +\mu_i^3} \text{.}
\end{align}
As $\E{\theta_i^2} = S_{ii} + \mu^2_i$, $\E{\theta_i^3}$ corresponds to the third non-central moment of the normal distribution, known to be $\mu_i^3 + 3\mu_i \S_{ii}$, the very last term in \eqref{eq:expectation_squared} simplifies to $\mu_i \S_{ii}$. Further noticing that $S_{ij} = S_{ji}$, 
\begin{align}\label{eq:expectation_term_1}
    \mathbb{E}[\theta_i \mathbb{E}[\theta_j^2\vert \theta_i]] &= \mu_i \sbr{S_{j\vert i} + \S_{ji}^2 \iS_{ii}}+2\mu_j \S_{ji}+ \mu_i \mu_j^2  \nonumber \\ 
    &= \mu_i\sbr{\S_{jj}-\S_{ji}\iS_{ii}\S_{ij}+\S_{ji}^2 \iS_{ii}}+2\mu_j \S_{ji}+ \mu_i \mu_j^2 \nonumber \\ 
    & = \mu_i\S_{jj}+2\mu_i \S_{ji}+\mu_i \mu_j^2  \text{.}
\end{align}
and by subtracting \eqref{eq:expectation_term_1} to \eqref{eq:expectation_term_2} as for \eqref{eq:cv_two_expectations}, we obtain
\begin{align}\label{eq:generic_cov_is_zero}
    \cov{\theta_j,\theta_i\theta_j} = \mu_i\S_{jj} + \mu_j\S_{ji} \text{.}
\end{align}
Thus, for the generic covariance term $\Cov\br{\theta_j,V_{ji}z_i}$ appearing in \eqref{eq:cov_theta_Vz},
\begin{align}\label{eq:cov_theta_Vz_ji}
     \Cov\br{\theta_j,V_{ji}z_i} &=z\br{\mu_i\S_{jj} + \mu_j\S_{ji}-\mu_i\S_{jj} - \mu_j\S_{ji}}
     =  0\text{,} \quad \forall i\neq j\text{.}
\end{align}
Therefore from \eqref{eq:cov_theta_Vz_jj} and \eqref{eq:cov_theta_Vz_ji}, the terms in \eqref{eq:cov_theta_Vz} are all zero: 
$$ \Cov(\theta_j,\br{Vz}_j) = 0 \text{,} \quad \forall j\text{.} $$
Returning to \eqref{eq:cov_grad_m1}, we finally have
\begin{align*}
    &\vcov\br{ \nabla_{m^{(1)}} \logq} = \iS \br{S + D} {\iS}\text{,}
\end{align*}
with $D = \vcov\br{Vz}$ given in \eqref{eq:kron_Vz}, which completes the proof.

\end{document}